\documentclass[10pt,twocolumn,letterpaper]{article}

\usepackage{cvpr}              

\usepackage{graphicx}
\usepackage{amsmath}
\usepackage{amssymb}
\usepackage{booktabs}
\usepackage{color}
\usepackage{multirow}
\usepackage{makecell}
\usepackage{bbm}
\usepackage{xcolor}         
\usepackage{soul}
\usepackage{newfloat}
\usepackage{listings}
\usepackage{diagbox}
\usepackage[square, comma, sort&compress, numbers]{natbib}
\usepackage[pagebackref,breaklinks,colorlinks]{hyperref}

\usepackage[capitalize]{cleveref}
\crefname{section}{Sec.}{Secs.}
\Crefname{section}{Section}{Sections}
\Crefname{table}{Table}{Tables}
\crefname{table}{Tab.}{Tabs.}

\makeatletter
\def\ps@myheadings{%
    \let\@oddfoot\@empty\let\@evenfoot\@empty
    \def\@evenhead{\thepage\hfil\slshape\leftmark}%
    \def\@oddhead{{\slshape\rightmark}\hfil\thepage}%
    \let\@mkboth\@gobbletwo
    \let\sectionmark\@gobble
    \let\subsectionmark\@gobble
    }
  \if@titlepage
  \renewcommand\maketitle{\begin{titlepage}%
  \let\footnotesize\small
  \let\footnoterule\relax
  \let \footnote \thanks
  \null\vfil
  \vskip 60\p@
  \begin{center}%
    {\LARGE \@title \par}%
    \vskip 3em%
    {\large
     \lineskip .75em%
      \begin{tabular}[t]{c}%
        \@author
      \end{tabular}\par}%
      \vskip 1.5em%
    {\large \@date \par}
  \end{center}\par
  \@thanks
  \vfil\null
  \end{titlepage}%
  \setcounter{footnote}{0}%
}
\else
\renewcommand\maketitle{\par
  \begingroup
    \renewcommand\thefootnote{\@fnsymbol\c@footnote}%
    \def\@makefnmark{\rlap{\@textsuperscript{\normalfont\color{black}\@thefnmark}}}%
    \long\def\@makefntext##1{\parindent 1em\noindent
            \hb@xt@1.8em{%
                \hss\@textsuperscript{\normalfont\@thefnmark}}##1}%
    \if@twocolumn
      \ifnum \col@number=\@ne
        \@maketitle
      \else
        \twocolumn[\@maketitle]%
      \fi
    \else
      \newpage
      \global\@topnum\z@   
      \@maketitle
    \fi
    \thispagestyle{plain}
    \@thanks
  \endgroup
  \setcounter{footnote}{0}%
}
\makeatother

\makeatletter
\newcommand\fs@nobottomruled{\def\@fs@cfont{\bfseries}\let\@fs@capt\floatc@ruled
  \def\@fs@pre{}
  \def\@fs@post{}
  \def\@fs@mid{\kern2pt\hrule\kern2pt}%
  \let\@fs@iftopcapt\iftrue}
\makeatother

\def\cvprPaperID{1853} 
\def\confName{CVPR}
\def\confYear{2023}

\begin{document}

\title{Self-supervised Implicit Glyph Attention for Text Recognition}

\author{Tongkun Guan\textsuperscript{\rm 1}, Chaochen Gu\textsuperscript{\rm 2\thanks{Corresponding author.}},
Jingzheng Tu\textsuperscript{\rm 2},
Xue Yang\textsuperscript{\rm 1},
Qi Feng\textsuperscript{\rm 2},
Yudi Zhao\textsuperscript{\rm 2},\\
Xiaokang Yang\textsuperscript{\rm 1},
Wei Shen\textsuperscript{\rm 1\footnotemark[1]}
\\
\textsuperscript{\rm 1} MoE Key Lab of Artificial Intelligence, AI Institute, Shanghai Jiao Tong University\\
\textsuperscript{\rm 2} Department of Automation, Shanghai Jiao Tong University\\
{\tt\small \{gtk0615,jacygu,wei.shen\}@sjtu.edu.cn}
}
\maketitle

\begin{abstract}
  The attention mechanism has become the \emph{de facto} module in scene text recognition (STR) methods, due to its capability of extracting character-level representations.
  These methods can be summarized into implicit attention based and supervised attention based, depended on how the attention is computed, i.e., implicit attention and supervised attention are learned from sequence-level text annotations and or character-level bounding box annotations, respectively. Implicit attention, as it may extract coarse or even incorrect spatial regions as character attention, is prone to suffering from an alignment-drifted issue. 
  Supervised attention can alleviate the above issue, but it is character category-specific, which requires extra laborious character-level bounding box annotations and would be memory-intensive when handling languages with larger character categories. 
  To address the aforementioned issues, we propose a novel attention mechanism for STR, self-supervised implicit glyph attention (SIGA). SIGA delineates the glyph structures of text images by jointly self-supervised text segmentation and implicit attention alignment, which serve as the supervision to improve attention correctness without extra character-level annotations. Experimental results demonstrate that SIGA performs consistently and significantly better than previous attention-based STR methods, in terms of both attention correctness and final recognition performance on publicly available context benchmarks and our contributed contextless benchmarks.
  Our code and two large-scale contextless datasets (MPSC and ArbitText) will be released in the future: \url{https://github.com/TongkunGuan/SIGA}. 
\end{abstract}
\section{Introduction}
\begin{figure}[t]
  \centering
  \graphicspath{{./graph/}}
  \includegraphics[width=3.2in]{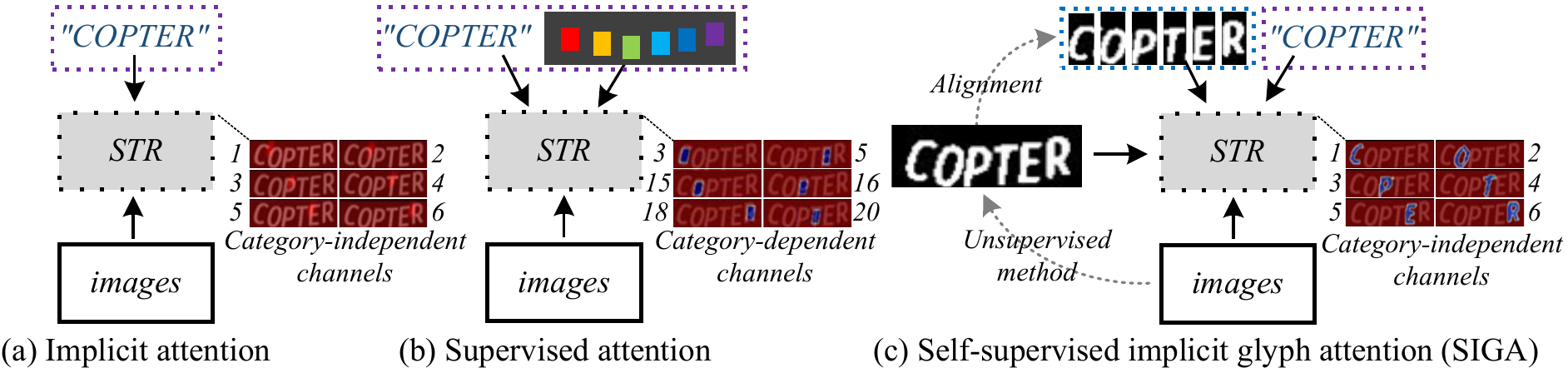}
  \vspace{-0.7em}
  \caption{Three difference supervised manners for STR.}
  \label{Figs.labels}
\end{figure}

Scene text recognition (STR) aims to recognize texts from natural images, which has wide applications in handwriting recognition~\cite{Graph-to-Graph,CoMER,PageNet}, industrial print recognition~\cite{guan2021industrial,9247163,10002724}, and visual understanding~\cite{biten2019scene,YXC002,script_identification}. Recently, attention-based models with encoder-decoder architectures are typically developed to address this task by attending to important regions of text images to extract character-level representations. 
These methods can be summarized into implicit attention methods (a) and supervised attention methods (b) as shown in Figure \ref{Figs.labels}, according to the annotation type used for supervising the attention.

Specifically, implicit attention is learned from sequence-level text annotations by computing attention scores across all locations over a 1D or 2D space. 
For example, the 1D sequential attention weights~\cite{ASTER:shi2018aster,TRBA:baek2019wrong} are generated at different decoding steps to extract important items of the encoded sequence. The 2D attention weights~\cite{SRN:yu2020towards,ABINET:fang2021read} are generated by executing a cross-attention operation with the embedded time-dependent sequences and visual features on all spatial locations. However, implicit attention methods, which only extract coarse or even unaligned spatial regions as character attention, may encounter alignment-drifted attention.
In contrast, supervised attention is learned from extra character-level bounding box annotations by generating character segmentation maps. Although these supervised attention methods~\cite{liao2019scene,Masktextspotter,wan2020textscanner,S_GTR} can alleviate the above issue, they rely on labour-intensive character-level bounding box annotations, and their attention maps with respect to character categories might be memory-intensive when the number of character categories is large. 

\begin{figure*}[t]
  \centering
  \graphicspath{{./graph/}}
  \includegraphics[width=6.8in]{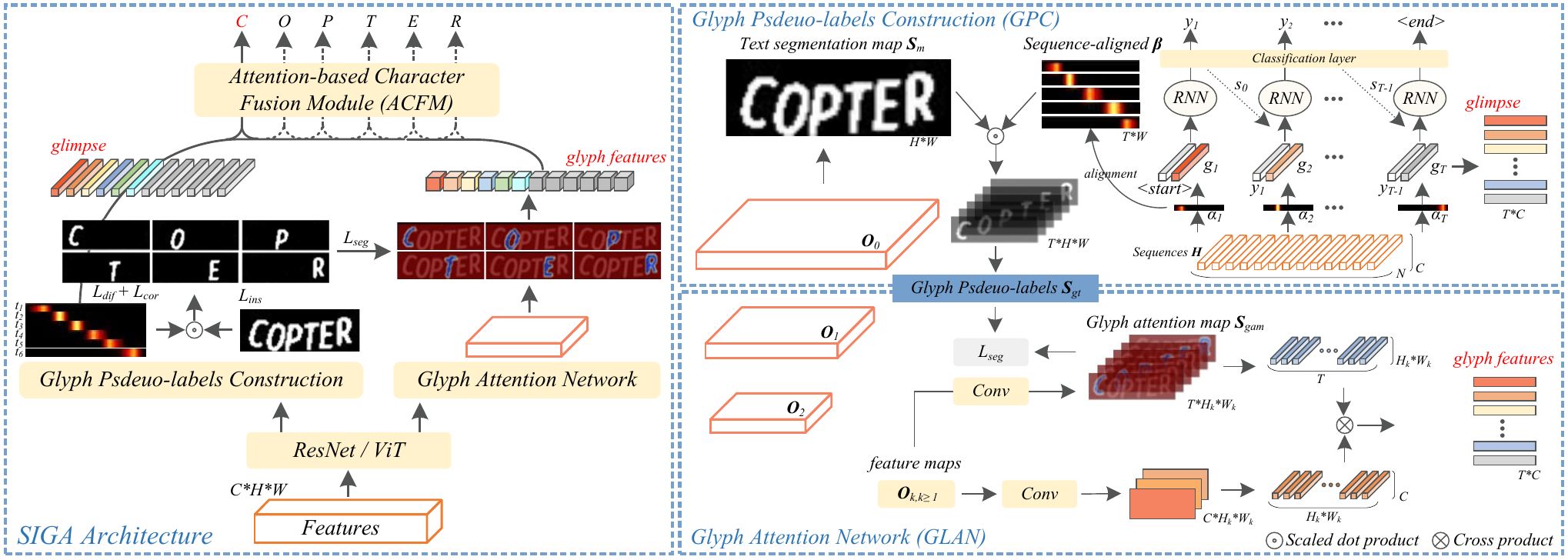}
  \vspace{-0.7em}
  \caption{Overview of the proposed self-supervised implicit glyph attention network (SIGA) for text recognition. 
  }
  \label{Figs.Network}
  \vspace{-1.2em}
\end{figure*}

To address the aforementioned issues, we propose a novel attention-based method for STR (Figure \ref{Figs.labels} (c)), self-supervised implicit glyph attention (SIGA). 
As briefly shown in Figure \ref{Figs.Network}, SIGA delineates the glyph structures of text images by jointly self-supervised text segmentation and implicit attention alignment, which serve as the supervision for learning attention maps during training to improve attention correctness.
Specifically, the glyph structures are generated by modulating the learned text foreground representations with sequence-aligned attention vectors. 
The text foreground representations are distilled from self-supervised segmentation results according to the internal structures of images~\cite{hartigan1979algorithm}; The sequence-aligned attention vectors are obtained by applying an orthogonal constraint to the 1D implicit attention vectors~\cite{TRBA:baek2019wrong}. They then serve as the position information of each character in a text image to modulate the text foreground representations to generate glyph pseudo-labels online.

By introducing glyph pseudo-labels as the supervision of attention maps, the learned glyph attention encourages the text recognition network to focus on the structural regions of glyphs to improve attention correctness.
Different from supervised attention methods, the glyph attention maps bring no additional cost to enable character and decoding order consistency when handling languages with larger character categories.

For recognizing texts with linguistic context, SIGA achieves state-of-the-art results on seven publicly available context benchmarks. We also encapsulate our glyph attention module as a plug-in component to other attention-based methods, achieving average performance gains of 5.68\% and 1.34\% on SRN~\cite{SRN:yu2020towards} and ABINet~\cite{ABINET:fang2021read}, respectively. 

It is worth mentioning that SIGA shows its prominent superiority in recognizing contextless texts widely used in industrial scenarios (\emph{e.g.}, workpiece serial numbers~\cite{guan2021industrial} and identification codes~\cite{9247163}).
Specifically, we contribute two large-scale contextless benchmarks (real-world MPSC and synthetic ArbitText) with random character sequences that differ from legal words. Experiments demonstrate that SIGA improves the accuracy of contextless text recognition by a large margin, which is 7.0\% and 10.3\% higher than MGP-STR~\cite{MGP} on MPSC and ArbitText, respectively. 
In summary, the main contributions are as follows:
\vspace{-0.5em}
\begin{itemize}
  \item We propose a novel attention mechanism for scene text recognition, SIGA, which is able to delineate glyph structures of text images by jointly self-supervised text segmentation and implicit attention alignment
  to improve attention correctness without character-level bounding box annotations.
  \item Extensive experiments demonstrate that the proposed glyph attention is essential for improving the performance of vision models. Our method achieves the state-of-the-art performance on publicly available context benchmarks and our contributed large-scale contextless benchmarks (MPSC and ArbitText).
\end{itemize}

\section{Related Work}
Recently, some top-down approaches have been developed to recognize entire images instead of directly recognizing character segments like traditional bottom-up approaches~\cite{Traditional_method:wang2010word, Traditional_method:neumann2012real}.
These methods can be roughly divided into language-free and language-aware methods.
\textbf{Language-free methods}
These methods view STR as a character-level classification task and mainly exploit the visual information to recognize texts.
According to the annotation types used for supervising the attention, implicit attention methods are developed for STR supervised by sequence-level text annotations, while supervised attention methods require additional character-level bounding box annotations.

Specifically, some 1D implicit attention methods~\cite{Focus_Attention:cheng2017focusing,EP:bai2018edit,Squeezedtext:liu2018squeezedtext,ASTER:shi2018aster,YXC001} execute sequence attention modeling over a 1D space. 
Input text images are first encoded into 1D sequential features. Then, they employ a bidirectional decoder to extract attentive features of the encoded sequence by outputting the corresponding attention weights for prediction.
Besides, some 2D implicit attention methods~\cite{SAR:li2019show,SATRN:lee2020recognizing,DAN:wang2020decoupled,SRN:yu2020towards,ABINET:fang2021read} develop various 2D attention mechanisms by attending to spatial vision features of each character of an image. 
For example, Li \emph{et al.}~\cite{SAR:li2019show} combine visual features with hidden states of the decoder to focus on spatial character features at each decoding step. 
Fang \emph{et al.}~\cite{ABINET:fang2021read} adopt a transformer-based structure to compute attention scores across all spatial locations of visual features, thereby obtaining attention maps of corresponding characters.
However, supervised by sequence-level text annotations, these implicit attention methods easily extract coarse or even unaligned spatial regions as character attention.

In contrast, under the supervision of extra character-level bounding box annotations, some supervised attention methods~\cite{Masktextspotter, wan2020textscanner, S_GTR} employ a fully convolutional network to predict character-level segmentation results and then perform classification tasks. For example, He \emph{et al.}~\cite{S_GTR} utilizes the segmentation probability maps to exploit spatial context for text reasoning by graph convolutional networks. However, character-level annotations of text images are expensive and laborious. Beyond the limitations of human annotations, we delineate the glyph structures of text images as the supervision of attention maps by jointly self-supervised text segmentation and implicit attention alignment.

\noindent \textbf{Language-aware methods}
Inspired by natural language processing methods~\cite{GPT,BERT:kenton2019bert}, the visual outputs of STR methods are fed into a language model to implement recognition correction with linguistic context. For example, some works~\cite{SRN:yu2020towards,Authors1} stack multiple layers of self-attention structures~\cite{transformer} for semantic reasoning tasks. 
Inspired by the masked language model (MLM) in BERT~\cite{BERT:kenton2019bert}, Fang \emph{et al.}~\cite{ABINET:fang2021read} pre-train the proposed BCN to predict the masked character in text based on linguistic context, and unite visual outputs to improve performance. Although these language-aware methods leverage a language model to optimize the joint character prediction probability with visual models, which reduces prediction errors with linguistic context, they do not generalize well to arbitrary texts (\emph{e.g.}, contextless texts with a random workpiece coding scheme).
Therefore extracting the distinctive visual features of characters is still the key to text recognition.
\vspace{-0.4em}
\section{Methodology}
\vspace{-0.3em}
In this section, we first review the representative attention-based method~\cite{TRBA:baek2019wrong} that implicitly learns the 1D attention weights, and then introduce our self-supervised implicit glyph attention method.
\subsection{Implicit Attention Method over 1D Space}
The implicit attention method~\cite{TRBA:baek2019wrong} consists of a transformation layer, an encoder and a decoder.
First, the transformation layer employs a Thin Plate Spline (TPS), a variant of the spatial transformation network (STN)~\cite{STN}, to transform an input image $\boldsymbol{X}$ into a normalized image $\boldsymbol{X'}$. 
Then, the encoder extracts sequential features $\boldsymbol{H}\in \mathbb{R}^{C\times 1 \times N}$ from the normalized image $\boldsymbol{X'}\in \mathbb{R}^{C\times H \times W}$ by a variant of ResNet~\cite{he2016deep}, and splits the sequential features into a fixed-length sequence $\{\boldsymbol{h}_{i}\}_{i={1,...,N}}$. In the decoder, as illustrated in Figure \ref{Figs.2} (a), the encoded sequence is fed into a recurrent module (\emph{e.g.}, LSTM, GRU) to generate an output vector $\boldsymbol{x}_{t}$ and a new state vector $\boldsymbol{s}_{t}$ at the decoding step $t$. The specific details are as follows:
\begin{gather}
  \label{eq0}
  (\boldsymbol{x}_{t}, \boldsymbol{s}_{t})={\rm rnn}(\boldsymbol{s}_{t-1}, (\boldsymbol{g}_{t}, E(y_{t-1}))), 
\end{gather}

\noindent where $(\boldsymbol{g}_{t}, E(y_{t-1}))$ denotes the combination of \emph{glimpse} $\boldsymbol{g}_{t}$ and the embedding vector of the predicted character category at the previous decoding step. 
Especially, $y_{0}$ denotes an artificially defined ``\textless start\textgreater'' token.
The \emph{glimpse} is computed by the attention mechanism as follows:
\begin{equation}
\label{eq3}
\begin{cases}
  \begin{aligned}
  e_{t,i} & = \boldsymbol{w}^{\intercal}{\rm tanh}(\boldsymbol{Ws}_{t-1}+\boldsymbol{Vh}_{i}+b),\\
  \alpha_{t,i} & = {\rm exp}(e_{t,i}) / \sum_{i' = 1}^{N}{\rm exp}(e_{t,i'}),\\
  \boldsymbol{g}_{t} & = \sum_{i}\alpha_{t,i}\boldsymbol{h}_{i}, i=1,...,N,
  \end{aligned}
\end{cases}
\end{equation}

\noindent where the $\boldsymbol{w,W,V}$ are learnable parameters. Finally, the output vector $\boldsymbol{x}_{t}$ predicts the character classification by a linear layer at the current decoding step t. 
The decoder is executed $T$ times (\emph{i.e.,} total decoding steps, $T$ = 26) and outputs classification results sequentially.
\subsection{Our Self-supervised Implicit Glyph Attention}
In this work, we follow the implicit attention method~\cite{TRBA:baek2019wrong} as the baseline structure, and delineate glyph structures of text images as the supervision of our attention network by proposing a novel online glyph pseudo-label construction module. The learned glyph attention encourages the text recognition network to focus on the structural regions of glyphs to improve attention correctness.
\subsubsection{Glyph Pseudo-label Construction (GPC)}
For a normalized image, given its text mask and the horizontal position information of each character, we can easily obtain the glyph structures of these characters by computing the dot product between them, instead of labour-intensive pixel-level annotations. Towards the goal, we construct glyph pseudo-labels online by jointly \textbf{\emph{self-supervised text segmentation}} and \textbf{\emph{implicit attention alignment}}. The sequence-aligned attentions serve as the position information of characters in a text image to modulate the learned text foreground representations to generate significant glyph pseudo-labels.

\begin{figure}[t]
  \centering
  \vspace{-0.8em}
  \graphicspath{{./graph/}}
  \includegraphics[width=3.2in]{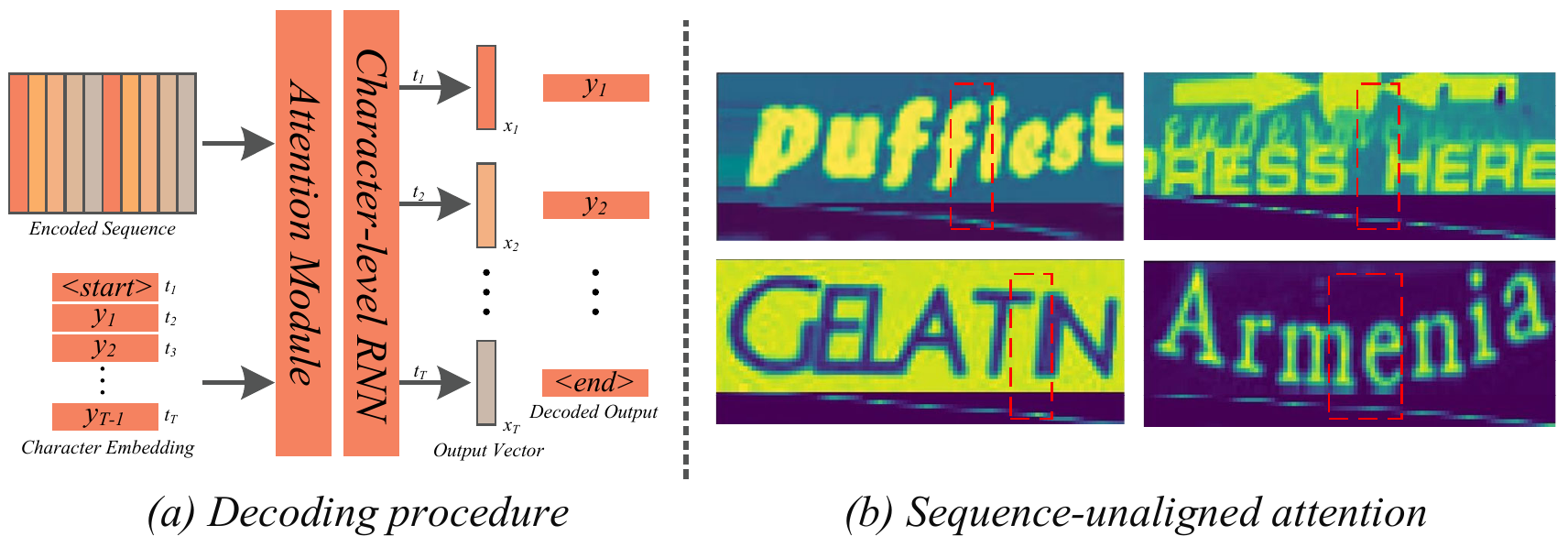}
  \vspace{-0.7em}
  \caption{Illustration of the representative attention-based decoder and some sequence-unaligned attention examples. The red dashed boxes in (b) indicate that the attention weight struggles to align text sequence at the current decoding step.}
  \label{Figs.2}
\end{figure}
\noindent \textbf{\emph{1) Self-supervised Text Segmentation.}}
In the subsection, we want to learn text foreground representations with morphological structures of glyphs, by a semantic segmentation network that assigns every pixel a foreground or background label on the unlabeled text images. 
It's observed that the underlying morphological representations of glyphs are not affected by slight structural changes (\emph{e.g.}, thicker or thinner), which reduces the reliance on pixel-level high-precision segmentation with expensive computation and annotation costs.
Inspired by the prior knowledge, we begin with a clustering task based on the internal structures of text images for obtaining pseudo-labels $\boldsymbol{S}_{\rm pl}$ about text masks. For simplicity, we focus our study on $K$-means to implement the clustering task, but other clustering approaches with predefined categories can be used. In the experiment, $K$ is set to 2, including the foreground and background categories.
Surprisingly, the morphological structures of glyphs are clustered well in most text images.

Then, the text foreground representations are distilled from self-supervised segmentation results produced by our designed text segmentation network. Specifically, we define the output of \emph{Conv 0}, \emph{Block 0}, and \emph{Block 1} from ResNet as $\boldsymbol{P}_0$, $\boldsymbol{P}_1$, and $\boldsymbol{P}_2$, and a top-down pyramid architecture is employed as follows:
\begin{equation}
  \label{eq4}
  \begin{cases}
    \begin{aligned}
      \boldsymbol{O}_2 &= \varphi(\boldsymbol{P}_2),\\
      \boldsymbol{O}_1 &= \varphi([\mathcal{T}(\boldsymbol{O}_2,s_1),\boldsymbol{P}_1]),\\
      \boldsymbol{O}_0 &= \varphi([\mathcal{T}(\boldsymbol{O}_1,s_0),\boldsymbol{P}_0]),
    \end{aligned}
  \end{cases}
  \end{equation}

\noindent where $\varphi(\cdot)$ denotes two convolutional layers with BatchNorm and ReLU activation function, 
$\mathcal{T}(\cdot)$ refers to a single $2 \times$ upsampling for $\boldsymbol{O}_k$ with resolution $s_k$ (\emph{i.e.}, $H_k \times W_k$), and $[\cdot]$ represents the concatenation operation along the channel axis. $\boldsymbol{O}_0$ is exploited to produce the text segmentation mask $\boldsymbol{S}_m$ by a binary classification convolutional layer.

\begin{figure}[t] 
  \centering
  \vspace{-0.5em}
  \graphicspath{{./graph/}}
  \includegraphics[width=3in]{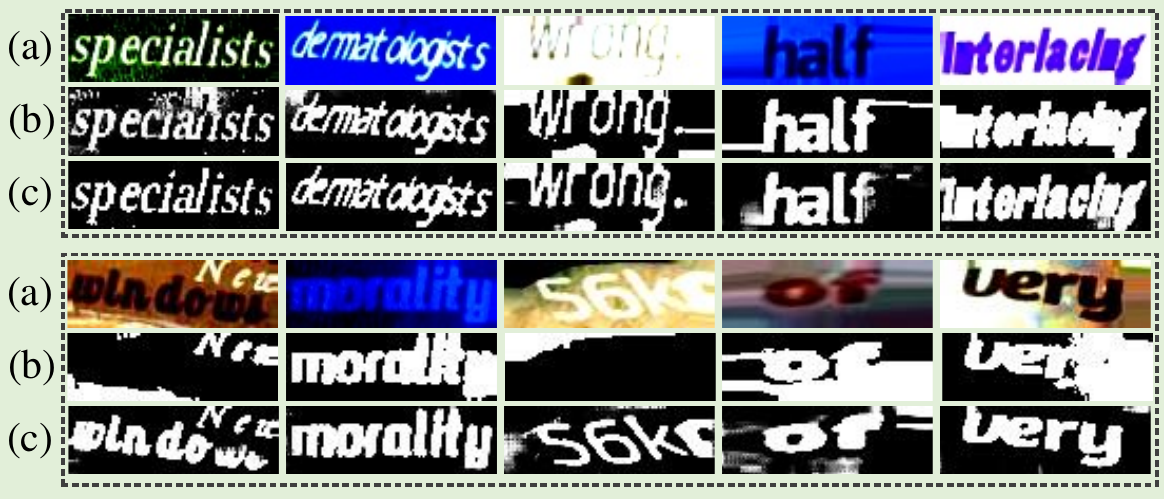}
  \vspace{-0.7em}
  \caption{Some segmentation examples. 
  (a) denotes the original images, (b) means the results of $K$-means, and (c) is our text segmentation maps. 
  In the last group of rows, the well-learned self-supervised text segmentation module can capture morphological structures of glyphs in challenging images.
  }
  \label{Figs.segmentation}
  \vspace{-1.2em}
\end{figure}

Finally, we employ a binary cross-entropy loss $\mathcal{L}_{\rm ins}$ between the text segmentation mask $\boldsymbol{S}_m$ and pseudo-labels $\boldsymbol{S}_{\rm pl}$ to optimize the text segmentation network. 
Consequently, the optimized segmentation network perceives the text foreground representations with morphological structures of glyphs in challenging text images, which may be difficult to be classified by an unsupervised clustering method $K$-means. Some visualization examples are shown in Figure \ref{Figs.segmentation}. 

\noindent \textbf{\emph{2) Implicit Attention Alignment.}} 
In the decoding unit (Eq. \ref{eq3}), the implicit attention weights $\boldsymbol{\alpha}$=$\{\boldsymbol{\alpha}_{t}\}_{t=1,...,T}$ focus on the important items of the encoded sequence to capture character dependencies. Inspiringly, we transform the attention weights as the position information of their corresponding characters. 
However, the time information of the decoder is drowned with the other introductions at the latter decoding steps, 
which easily leads to alignment drift as shown in Figure \ref{Figs.2} (b), \emph{i.e.}, the learnable attention weights struggle to align the text sequence~\cite{DAN:wang2020decoupled}.

To address the issue, we apply an orthogonal constraint to the implicit attention weights to obtain sequence-aligned attention vectors.
Specifically, we take these learnable attention weights as vectors and perform an alignment operation by ensuring that they are orthogonal to each other and that each processed vector is aligned with the corresponding character of the text segmentation mask $\boldsymbol{S}_m$. 
Assuming that $L$ denotes the character number of a text image, we first calculate the correlation coefficient $S_{\rm cor}$ between $L$ attention vectors, and then extract the character saliency map $\boldsymbol{S}_{\rm sal}$ by the attention vectors. 
The details are as follows:
\begin{equation}
  \begin{cases}
    \begin{aligned}
      S_{\rm cor} & = \sum_{1\leqslant t \textless t' \leqslant L} \boldsymbol{\alpha}^\intercal_{t}\boldsymbol{\alpha}_{t'},\\
      \boldsymbol{\beta}_{t} &= \xi(\boldsymbol{\alpha}_{t}),\\
      \boldsymbol{S}_{\rm sal} & = \sum_{t=1}^{L}(\sigma(\boldsymbol{\beta}_{t})\cdot\boldsymbol{S}_m),
    \end{aligned}
  \end{cases}
\end{equation}

\noindent where $\xi$ represents the one-dimensional linear interpolation ($\xi: \boldsymbol{\alpha}_{t} \in \mathbb{R}^{N} \rightarrow \boldsymbol{\beta}_{t} \in \mathbb{R}^{W}$). $\sigma(\cdot)$ refers to the nonlinear activation function, which maps each element in the vector to [0,1]:
{\setlength\abovedisplayskip{1pt}
\setlength\belowdisplayskip{1pt}
\begin{equation}
  \label{eq1}
  \sigma(x) = 1/(1+{\rm exp}(-\mu (x-\lambda ))),
\end{equation}}

\noindent where $\mu,\lambda$ represent scaling and offset transitions, set to 70 and 0.1 in the experiment, respectively.   
Then the alignment drift problem can be alleviated by minimizing the correlation coefficient $S_{\rm cor}$ and the difference $S_{\rm dif}$ between $\boldsymbol{S}_m$ and $\boldsymbol{S}_{\rm sal}$ by the following loss function:
{\setlength\abovedisplayskip{1pt}
\setlength\belowdisplayskip{1pt}
\begin{equation}
  \begin{cases}
    \begin{aligned}
      \mathcal{L}_{\rm cor} &= S_{\rm cor},\\
      \mathcal{L}_{\rm dif} &= \frac{1}{n}\sum_{i=1}^{n}-(\rho_{i}\log(\rho_{i}^{*}) + (1-\rho_{i})\log(1-\rho_{i}^{*})),\\
      \mathcal{L}_{\rm seq} &= \mathcal{L}_{\rm cor} + \mathcal{L}_{\rm dif},
    \end{aligned}
  \end{cases}
\end{equation}}

\noindent where $n$ denotes the number of pixels in the text segmentation map $\boldsymbol{S}_m$, $\rho_{i}$ and $\rho_{i}^{*}$ are the confidence score of pixel $i$ in $\boldsymbol{S}_m$ and $\boldsymbol{S}_{\rm sal}$, respectively.

Finally, by optimizing the proposed constraint function during training, the attention weights are successfully aligned with the encoded sequence and contribute accurate positional information for glyph pseudo-label construction.

\begin{figure}[t]
  \centering
  \graphicspath{{./graph/}}
  \includegraphics[width=3in]{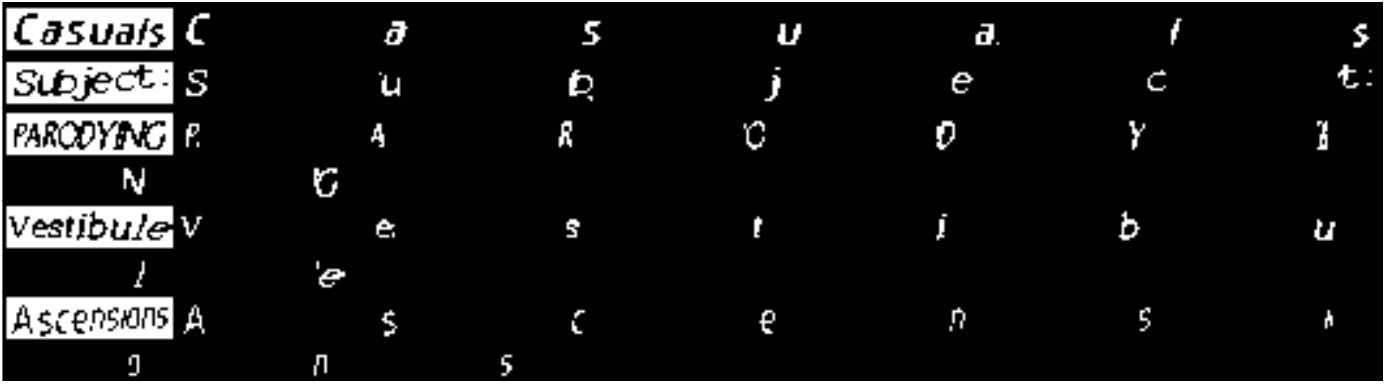}
  \vspace{-0.7em}
  \caption{Self-constructed glyph pseudo-labels online.}
  \label{Figs.gt}
  \vspace{-1.2em}
\end{figure}

\noindent \textbf{\emph{3) Glyph Pseudo-label Construction.}}
By calculating the dot product between the aligned attention weights $\boldsymbol{\beta}$=$\{\boldsymbol{\beta}_{t}\}_{t=1,...,T}$ and text segmentation mask $\boldsymbol{S}_m$, we obtain the glyph pseudo-labels.
Specifically, assuming that the glyph pseudo-label is $\boldsymbol{S}_{\rm gt}$, we construct it from the concatenation operation as follows:
{\setlength\abovedisplayskip{0cm}
\setlength\belowdisplayskip{0.1cm}
\begin{equation}
  \label{eq2}
  \boldsymbol{S}_{\rm gt} = [1-\boldsymbol{S}_m,\mathbbm{1}_{[\boldsymbol{\beta}_{1} \geqslant \delta]} \cdot \boldsymbol{S}_m,...,\mathbbm{1}_{[\boldsymbol{\beta}_{T} \geqslant \delta]} \cdot \boldsymbol{S}_m],
\end{equation}}

\noindent where $[\cdot]$ represents the concatenation operation along the channel axis. $\delta$ denotes the confidence threshold, which is set to 0.05 in the experiment. Some visualization examples about the self-constructed $\boldsymbol{S}_{\rm gt}$ are shown in Figure \ref{Figs.gt}.  

Note that the proposed glyph pseudo-label construction module will be removed in the test stage.
\vspace{-0.5em}
\subsubsection{Glyph Attention Network (GLAN)}
The existing supervised attention methods for STR have the following limitations: 
1) For languages with larger character categories, these methods might be memory-intensive and run slower due to their category-dependence character segmentation maps.
2) It is not easy to obtain the character order of text directly from the character segmentation maps predicted by CNNs. An extra order segmentation branch is usually introduced to ensure channel and decoding order consistency, which brings time and computational complexity.
3) Training the segmentation network requires laborious and difficult character-level bounding box annotations. 

Benefiting from self-constructed glyph pseudo-labels, our glyph attention network does not have these limitations since the order is ensured and characters are well aligned. The glyph attention network generates glyph attention maps with fixed-length and category-independent channels, whose channel-specific map corresponds to the order-specific glyph attention. 
Specifically, followed by several convolutional layers, the features $\boldsymbol{O}_{k}$ in Eq. \ref{eq4} are utilized to predict a glyph attention map $\boldsymbol{S}_{\rm gam}$ with a channel number of $N_{s}$. 
$N_{s}$ is set to $1+M$ and not equal to the character categories, which represents the sum of the background category and the set maximum character length on text images ($M$ = 26).
For example, to recognize GB2312 with 6763 categories in Chinese, if the same convolution layer is employed and feature channels are 256, the parameter size is 1.7M (256$\times$6763) for supervised attention methods while 6.9K (256$\times$27) for our method.

And then, supervised by the constructed glyph pseudo-labels $\boldsymbol{S}_{\rm gt}$, we use the joint loss function of multi-class Dice loss~\cite{UNET} and cross-entropy loss to boost the segmentation performance of the glyph attention network. The specific details are as follows:
\vspace{-0.6em}
\begin{equation}
\small
  \begin{cases}
    \begin{aligned}
      \mathcal{L}_{\rm dice} &=\frac{1}{L}\sum_{j=2}^{L+1} (1-\frac{2\sum_{i=1}^{n}(\omega_{j,i}\omega_{j,i}^{*})}{\sum_{i=1}^{n}(\omega_{j,i}) + \sum_{i=1}^{n}(\omega_{j,i}^{*})}),\\
      \mathcal{L}_{\rm cel} &= \frac{-1}{n}\sum_{i=1}^{n}(\rho_{i}\log(\sum_{j=2}^{M+1}\omega_{j,i}^{*}) + (1-\rho_{i})\log(1-\sum_{j=2}^{M+1}\omega_{j,i}^{*})),\\
      \mathcal{L}_{\rm seg} &= \mathcal{L}_{\rm dice} + \mathcal{L}_{\rm cel},
    \end{aligned}
  \end{cases}
\end{equation}
where 
$\omega_{j,i}$ and $\omega_{j,i}^{*}$ are the confidence scores of the $i$-th pixel $p_{i}$ of the $j$-index map in the pseudo-label $\boldsymbol{S}_{\rm gt}$ and glyph attention map $\boldsymbol{S}_{\rm gam}$, respectively. $\rho_{i}$ is the confidence score of the $p_{i}$ in $\boldsymbol{S}_m$. $L$ denotes the character number of a text image.

Finally, the learned glyph attention encourages the recognition branch to focus on the structural regions of glyphs to extract glyph features for STR, which contain more robust and discerning character representations. 
Specifically, the encoded text features $\boldsymbol{O}_{k} \in \mathbb{R}^{W_k\times H_k\times C}$ are first fed into two convolutional layers with BatchNorm and ReLU activation functions, and then multiplied with glyph attention maps $\boldsymbol{S}_{\rm gam} \in \mathbb{R}^{W_k\times H_k\times M}$ (remove background) to obtain glyph features $\boldsymbol{I}_{k} \in \mathbb{R}^{M\times C}$. 
\begin{table}[t]
  \setlength{\tabcolsep}{6pt}
  \centering 
  \caption{The parameter setting table of SIGA.}
  \vspace{-0.8em}
  \label{table:parameter}
  \scalebox{0.8}{
  \begin{tabular}{ll|ll}
  \hline\noalign{\smallskip}
  Name & Value & Name & Value\\
  \noalign{\smallskip}
  \hline
  \noalign{\smallskip}
  Decoding step $T$ & 26 & Sequence length $N$ & 32 \\
  Feature channel $C$ & 256 & Max character length $M$ & 26 \\
  Image size $W,H$ & 128, 32 & Constant $\lambda,\mu $ & 0.1, 70 \\
  Confidence threshold $\delta $ & 0.05 & Constant $k $ & 2 \\
  \hline
  \end{tabular}
  }
  \vspace{-1.2em}
\end{table}
\begin{table*}[t]
  \centering
  \caption{Comparison results of language-free STR methods. 
  $\dagger$ represents the visual model performance for a fair comparison. $^*$ combines 21 mixing blocks (10 local blocks and 11 global blocks) with local and global modeling capabilities for extracting features. ``Trns'' refers to several transformer units~\cite{transformer} consisting of a MHSA and a FFN. ``SATRN" is tailored for a transformer-based text feature extractor~\cite{SATRN:lee2020recognizing}.
  These symbols follow the same convention within the scope of this paper. The best results are shown in bold font. Underline values represent the second-best results. 
  }
  \vspace{-0.8em}
  \scalebox{0.6}{
  \begin{tabular}{c|c|c|c|c|p{7mm}<{\centering}|p{7mm}<{\centering}|p{7mm}<{\centering}p{7mm}<{\centering}|p{7mm}<{\centering}p{7mm}<{\centering}|p{7mm}<{\centering}p{7mm}<{\centering}|p{7mm}<{\centering}|p{7mm}<{\centering}c}
  \toprule
  \multirow{2}{*}{Methods} &\multirow{2}{*}{Backbone}&\multirow{2}{*}{Structure} &\multirow{2}{*}{Size} & \multirow{2}{*}{Venue} & IIIT & SVT & \multicolumn{2}{c|}{IC03} & \multicolumn{2}{c|}{IC13} & \multicolumn{2}{c|}{IC15} & SP & CT \\
  & && &  & 3000 & 647 & 860    & 867   & 857  & 1015 & 1811  & 2077   & 645 & 288 &\\ 
  \midrule
  CA-FCN~\cite{liao2019scene}  &VGG16&\multirow{13}{*}{CNN} &64$\times$256  &AAAI2019  &91.9 &86.4 &- &- &- &91.5 &-  &- &-&79.9 &\\
  DAN~\cite{DAN:wang2020decoupled} &ResNet45& &32$\times$128 &AAAI2020 &94.3 &89.2 &- &95.0 &- &93.9 &- &74.5 &80.0 &84.4 &\\
  TextScanner~\cite{wan2020textscanner} &ResNet50& &64$\times$256 &AAAI2020 &93.9 &90.1 &- &- &- &92.9 &79.4 &- &84.3 &83.3&\\
  SRN$\dagger$~\cite{SRN:yu2020towards} &ResNet50-FPN& &64$\times$256 &CVPR2020 &92.3 &88.1 &- &- &- &93.2 &77.5 &- &79.4 &84.7&\\
  PlugNet~\cite{mou2020plugnet} &ResNet37& &32$\times$100 &ECCV2020 &94.4 &92.3 &95.7 &- &- &95.0 &- &\textbf{82.2} &84.3 &85.0&\\
  PIMNet~\cite{PIMNet} &ResNet50-FPN& &64$\times$256  &ACM MM2021 &95.2 &91.2 &- &- &95.2 &93.4 &83.5 &81.0 &84.3 &84.4&\\
  TRBA~\cite{TRBA:baek2019wrong} &ResNet31& &32$\times$100  &CVPR2021 &92.1 &88.9 &94.8 &\underline{95.1} &93.9 &93.1 &78.3 &74.7 &79.5 &78.2&\\
  PREN2D~\cite{yan2021primitive} &EfficientNet-B3& &64$\times$256  &CVPR2021 &\underline{95.6} &\textbf{94.0} &\underline{95.8} &- &\underline{96.4} &- &83.0 &- &\textbf{87.6} &\textbf{91.7}&\\
  Text is Text~\cite{bhunia2021text} &ResNet31& &48$\times$160 &ICCV2021 &92.3 &89.9 &- &- &93.3 &- &- &76.9 &84.4 &86.3&\\
  S-GTR$\dagger$~\cite{S_GTR} &ResNet50Dilated-PPM& &64$\times$256 &AAAI2022 &94.0 &91.2 &- &- &94.8 &- &82.8 &- &85.0 &\underline{88.4}&\\
  LevOCR$\dagger$~\cite{LevOCR} &ResNet45 & &32$\times$128 &ECCV2022 &95.2 &90.6 &-&- &95.1 &- &\underline{84.0} &-  &83.4 &87.8 &\\
  SGBANet~\cite{zhong2022sgbanet} &ResNet45-FPN& &64$\times$256 &ECCV2022 &95.4 &89.1 &- &- &- &\underline{95.1} &- &78.4 &83.1 &88.2&\\
  SIGA$_{R}$ (ours) &ResNet45& &32$\times$128 &-&\textbf{95.9} &\underline{92.7} &\textbf{96.5} &\textbf{95.9} &\textbf{97.0} &\textbf{95.6} &\textbf{85.1} &\underline{81.7} &\underline{87.1} &\textbf{91.7}&\\
  \midrule
  SVTR~\cite{SVTR} &SVTR-L&\multirow{2}{*}{Transformer$^*$} &48$\times$160 &IJCAI2022 &\underline{96.3} &\underline{91.7} &- &- &\textbf{97.2} &- &\underline{86.6} &- &\underline{88.4} &\textbf{95.1}&\\
  SIGA$_{S}$ (ours) &SVTR-L& &48$\times$160 &-&\textbf{96.9} &\textbf{93.7} &- &- &\underline{97.0} &- &\textbf{87.6} &- &\textbf{89.5} &\underline{92.0}&\\
  \midrule
  ViTSTR~\cite{VITSTR:atienza2021vision} &ViT-B&\multirow{7}{*}{Transformer} &224$\times$224 &ICDAR2021 &88.4 &87.7 &\underline{94.7} &94.3 &93.2 &92.4 &78.5 &72.6  &81.8 &81.3&\\
  ABINet$\dagger$~\cite{ABINET:fang2021read}  &ResNet45-Trns& &32$\times$128  &CVPR2021 &94.7 &91.7 &94.3 &\underline{94.7} &95.0 &93.6 &82.7 &\underline{83.0} &85.1 &86.5&\\
  ABINet+ConCLR$\dagger$~\cite{ConCLR}  &ResNet45-Trns& &32$\times$128  &AAAI2022 &95.7 &92.1 &- &- &- &95.9 &84.4 &-  &85.7 &89.2&\\
  LevOCR$\dagger$~\cite{LevOCR} &ViT & &32$\times$128 &ECCV2022 &93.6 &89.2 &-&- &94.9 &- &82.4 &-  &84.2 &83.0 &\\
  CornerTransformer~\cite{xie2022toward}  &SATRN& &32$\times$128  &ECCV2022 &95.9 &94.6 &- &- &- &\underline{96.4} &- &\textbf{86.3}  &\textbf{91.5} &\underline{92.0}&\\
  MGP-STR~\cite{MGP}  &ViT-B& &32$\times$128  &ECCV2022 &\underline{96.4} &\underline{94.7} &- &- &\underline{97.3} &- &\textbf{87.2} &- &\underline{91.0} &90.3&\\
  SIGA$_{T}$ (ours) &ViT-B& &32$\times$128  &- &\textbf{96.6} &\textbf{95.1} &\textbf{96.9} &\textbf{97.0} &\textbf{97.8} &\textbf{96.8} &\underline{86.6} &\underline{83.0} &90.5 &\textbf{93.1}&\\
  \bottomrule
\end{tabular}
}
  \label{tb:results}
  \vspace{-0.5em}
\end{table*}
\vspace{-0.5em}
\begin{table*}[t]
  \centering
  \caption{Comparison results of language-aware STR methods. 
  ``V'' and ``VL'' types refer to language-free model and language-aware model, respectively.
  The best results are shown in bold font. Underline values represent the second-best results.
  }
  \vspace{-0.8em}
  \scalebox{0.6}{
  \begin{tabular}{c|c|c|c|c|c|p{7mm}<{\centering}|p{7mm}<{\centering}|p{7mm}<{\centering}p{7mm}<{\centering}|p{7mm}<{\centering}p{7mm}<{\centering}|p{7mm}<{\centering}|p{7mm}<{\centering}c}
  \toprule
  \multirow{2}{*}{Methods} &\multirow{2}{*}{Types} &\multirow{2}{*}{Backbone} &\multirow{2}{*}{Structure}&\multirow{2}{*}{Size} & \multirow{2}{*}{Venue} & IIIT & SVT & \multicolumn{2}{c|}{IC13} & \multicolumn{2}{c|}{IC15} & SP & CT &\\
  & & & &&  & 3000 & 647       & 857        & 1015        & 1811        & 2077          & 645       & 288 &\\ 
  \midrule
  SRN~\cite{SRN:yu2020towards} &VL &ResNet50-FPN&\multirow{6}{*}{ResNet} &64$\times$256  &CVPR2020  &94.8 &91.5  &- &95.5 &82.7 &- &85.1  &87.8 &\\
  Bhunia et al.~\cite{Authors1} &VL &ResNet50-FPN& &32$\times$100 &ICCV2021  &95.2 &92.2  &- &95.5 &- &\textbf{84.0} &85.7 &89.7&\\
  VisionLAN~\cite{visionLAN} &VL &ResNet45& &64$\times$256  &ICCV2021   &95.8 &91.7 &- &\textbf{95.7} &83.7 &- &86.0 &88.5&\\
  S-GTR~\cite{S_GTR} &VL &ResNet50Dilated-PPM& &64$\times$256 &AAAI2022 &95.8 &\textbf{94.1} &96.8 &- &84.6 &- &\underline{87.9} &\textbf{92.3}&\\
  LevOCR~\cite{LevOCR} &VL &ResNet45& &32$\times$128 &ECCV2022 &\textbf{96.6} &\underline{92.9} &\underline{96.9} &- &\textbf{86.4} &-  &\textbf{88.1} &\underline{91.7} &\\
  SIGA$_{R}$ (ours) &V &ResNet45& &32$\times$128  &-&\underline{95.9} &92.7  &\textbf{97.0} &\underline{95.6} &\underline{85.1} &\underline{81.7} &87.1 &\underline{91.7}&\\
  \midrule
  ABINet~\cite{ABINET:fang2021read} &VL &ResNet45-Trns&\multirow{5}{*}{Transformer} &32$\times$128 &CVPR2021  &96.2 &93.5  &\underline{97.4} &- &86.0 &- &\underline{89.3} &89.2&\\
  ABINet+ConCLR~\cite{ConCLR}  & VL &ResNet45-Trns& &32$\times$128  &AAAI2022 &96.5 &94.3 &- &\textbf{97.7} &85.4 &-  &\underline{89.3} &91.3 &\\
  PARSeq~\cite{bautista2022parseq} &VL &DeiT& &32$\times$128 &ECCV2022 &\textbf{97.0} &\underline{93.6} &97.0 &96.2 &\underline{86.5} &\underline{82.9}  &88.9 &\underline{92.2} &\\
  LevOCR~\cite{LevOCR} &VL &ViT& &32$\times$128 &ECCV2022 &95.6 &91.8 &96.2 &- &85.8 &-  &88.1 &86.8 &\\
  SIGA$_{T}$ (ours) &V &ViT-B& &32$\times$128  &- &\underline{96.6} &\textbf{95.1} &\textbf{97.8} &\underline{96.8} &\textbf{86.6} &\textbf{83.0} &\textbf{90.5} &\textbf{93.1} &\\
  \bottomrule
\end{tabular}
}
  \label{tb:language}
  \vspace{-0.9em}
\end{table*}
\subsubsection{Attention-based Character Fusion Module}
As discussed above, the visually aligned \emph{glimpse} $\boldsymbol{g}_{t}$ and glyph features $\boldsymbol{I}_{k,t}$ denote two different character feature representations at the decoding step $t$. Considering that their contributions to STR should be different among various text images, inspired by the gate unit~\cite{Gate}, we dynamically fuse the sequence $\boldsymbol{I}_{k,t}$ and \emph{glimpse} $\boldsymbol{g}_{t}$ to enrich the semantic information for character recognition.
Finally, we embed the final sequence into a decoder~\cite{TRBA:baek2019wrong} to output the current decoded classification result. 
\section{Experiments}
\subsection{Datasets} 
\begin{figure}[t]
  \centering
  \graphicspath{{./graph/}}
  \includegraphics[width=3.2in]{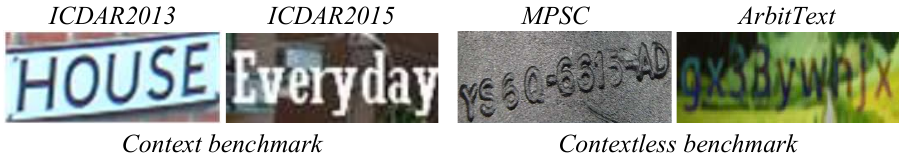}
  \vspace{-0.7em}
  \caption{Comparison of different text benchmarks of STR.}
  \vspace{-1.2em}
  \label{Figs.0}
\end{figure}

Our model is trained on two large-scale synthetic datasets (\emph{i.e.} SynthText~\cite{ST:gupta2016synthetic} and MJSynth~\cite{MJ:jaderberg2014synthetic}) for a fair comparison. 
Nine STR datasets are used to evaluate the performance of our method, including seven publicly available context benchmarks~\cite{TRBA:baek2019wrong} (\emph{i.e.}, IIIT5K-Words, ICDAR2003, ICDAR2013, Street View Text, ICDAR2015, SVT Perspective, and CUTE80) and two contextless benchmarks (MPSC and ArbitText). The differences between context and contextless benchmarks are shown in Figure \ref{Figs.0}.

\noindent \textbf{MPSC:} We cropped 15003 real-world text instances from industrial images marked workpiece information~\cite{guan2021industrial}, which is larger than the sum of seven context benchmarks. These texts are randomly collected from massive internet images and not from the same batch of products, containing various workpieces with irregular character combinations (\emph{e.g.}, ``YS6Q-6615-AD", ``TBJU8549728", and ``RS550SH-4941") for marking workpiece information. 

\noindent \textbf{ArbitText:} We also synthesize a contextless ArbitText with 1M images, and every sample is generated by a random combination of English letters and Arabic numerals.
\subsection{Implementation Details}
The parameter details are shown in Table \ref{table:parameter}. 
According to the backbone types of the existing text recognition models, we construct three typical SIGA architectures, \emph{i.e.}, SIGA$_{R}$, SIGA$_{S}$ and SIGA$_{T}$, for a fair comparison. 
For SIGA$_{R}$ with a ResNet45~\cite{he2016deep} as the backbone, we adopt the Adam optimizer~\cite{kingma2014adam} and the one-cycle learning rate scheduler~\cite{One_cycle} with a maximum learning rate of 0.0005 to train our model. 
We employ the same augmentation strategy from ABINet~\cite{ABINET:fang2021read} and set the batch size to 512 and the training epoch to 6.
For SIGA$_{S}$, we select SVTR-L~\cite{SVTR} with local and global modeling capabilities as the backbone, and the training parameters are the same as SIGA$_{R}$.
For SIGA$_{T}$ using ViT~\cite{ViT} as the backbone, we utilize the same setting including optimizer, learning rate scheduler, and batch size from MGP-STR~\cite{MGP}. 
To adapt the transformer structure to our method, we select the output of 2, 4, and 6 layers of ViT as $\boldsymbol{P}_0$, $\boldsymbol{P}_1$, and $\boldsymbol{P}_2$ to execute the GPC module. 
\subsection{Comparisons on context benchmarks}
\noindent \textbf{Language-free model.} The language-free methods mainly exploit visual information to recognize texts. As shown in Table \ref{tb:results}, we compare with the previous state-of-the-art language-free methods according to backbone types to fairly evaluate the effectiveness of our method on standard context benchmarks. 

For the CNN-based methods, SIGA$_{R}$ achieves state-of-the-art performance on seven context benchmarks.
Specifically, compared to supervised attention methods 
(CA-FCN~\cite{liao2019scene} and TextScanner~\cite{wan2020textscanner}), SIGA$_{R}$ doesn't need extra character-level annotations and brings significant performance gains (2.0\% $\sim$ 11.8\%) on these benchmarks. Compared to implicit attention methods, SIGA$_{R}$ has better performances and outperforms the second-best results on IIIT, IC03-860, IC03-867, IC13-857, IC13-1015, and IC15-1811 benchmarks by 0.3\%, 0.7\%, 0.8\%, 0.6\%, 0.5\%, and 1.1\%, respectively. 
SIGA$_{R}$ also achieves competitive performances (underline values) on SVT, IC15-2077, and SP benchmarks.

We also deploy the backbone of SVTR~\cite{SVTR} to implement STR, and consequently, SIGA$_{S}$ gets higher accuracy on four of its reported six standard benchmarks, with an average accuracy improvement of 0.63\%.

For the Transformer-based methods, SIGA$_{T}$ shows its prominent superiority and achieves state-of-the-art results on IIIT, SVT, IC03-860, IC03-867, IC13-857, IC13-1015, and CT benchmarks. Besides, we also obtain competitive results on IC15-1811 and IC15-2077 benchmarks.
These results demonstrate the effectiveness of our method on context benchmarks, as more discerning visual features are successfully extracted by introducing glyph attention.

\noindent \textbf{Language-aware model.} 
The semantic reasoning task of language models corrects visual outputs to reduce prediction errors with linguistic context (\emph{e.g.}, correcting ``unjversity" to ``university"), which improves the overall recognition accuracy on context benchmarks. 
As shown in Table \ref{tb:language}, when further compared with these language-aware methods, SIGA$_{R}$ achieves competitive results on the most standard benchmarks, and SIGA$_{T}$ gets the best accuracy on six of the eight benchmarks despite not using the semantic reasoning task on benchmarks with linguistic context. Specifically, SIGA$_{T}$ has better performances on SVT, IC13-857, SP and CT benchmarks by 1.5\%, 0.4\%, 1.2\%, and 0.9\%, respectively, which implies that a visual model could still perform well on context benchmarks. 

\subsection{Comparisons on contextless benchmarks}
Contextless texts consist of random character sequences, which are widely used in industrial scenarios~\cite{guan2021industrial,9247163}. Different from context benchmarks, they contain less semantic information. 
Thus language-aware methods that build implicit language representations with linguistic context are unsuitable for these contextless texts. Exploiting the visual features of text images is crucial for improving recognition accuracy on contextless benchmarks. As shown in Table \ref{tb:contextless}, we conduct comparative experiments with other language-free text recognition methods by utilizing the same training data (MJ and ST). These results are obtained by directly loading their released checkpoints to be evaluated. Specifically, we first evaluate on our contributed real-world MPSC benchmark. Then, we also synthesize a large-scale contextless benchmark with 1M text images, ArbitText, to further evaluate the generality and effectiveness of language-free models.

Consequently, our method shows a significant superiority on these contextless benchmarks, as SIGA$_{R}$ is 4.8\% and 4.9\% higher than PIMNet~\cite{PIMNet}, SIGA$_{S}$ is 1.3\% and 2.9\% higher than SVTR~\cite{SVTR}, SIGA$_{T}$ is 7.0\% and 10.3\% higher than MGP-STR~\cite{MGP} on the real-world MPSC and synthetic ArbitText benchmarks.


These results consistently emphasize that SIGA can generalize well to arbitrary texts (context benchmarks and contextless benchmarks), and the proposed glyph attention is essential for improving the performance of visual models.
\begin{table}[t]
  \centering
  \caption{Comparison results of contextless benchmarks.}
  \vspace{-0.8em}
  \scalebox{0.7}{
  \begin{tabular}{c|c|c|c}
  \toprule
  \multirow{2}{*}{Methods} &\multirow{2}{*}{Venue} & MPSC & ArbitText  \\
    && 15003 & 1000000 \\ 
  \midrule
  SAR~\cite{SAR:li2019show} &AAAI2019 &59.7 &64.5 \\
  DAN~\cite{DAN:wang2020decoupled} &AAAI2020 &57.7 &61.0 \\
  GA-SPIN~\cite{SPIN} &AAAI2021 &51.7 &54.0 \\
  PIMNet~\cite{PIMNet} &ACM MM2021 &60.8 &61.1\\
  SIGA$_{R}$ (ours) &- &\textbf{65.6} &\textbf{66.0} \\
  \midrule
  SVTR~\cite{SVTR} &IJCAI2022 &71.4 &78.1\\
  SIGA$_{S}$ (ours) &- &\textbf{72.7} &\textbf{81.0} \\
  \midrule
  ABINet$\dagger $~\cite{ABINET:fang2021read} &CVPR2021 & 64.4 &61.8 \\
  MGP-STR~\cite{MGP} &ECCV2022 &65.0 & 61.4 \\
  SIGA$_{T}$ (ours) &- &\textbf{72.0} &\textbf{71.7} \\
  \bottomrule    
  \end{tabular}
  }
  \label{tb:contextless}
  \end{table}
\begin{figure}[t]
  \centering
  \graphicspath{{./graph/}}
  \includegraphics[width=3.2in]{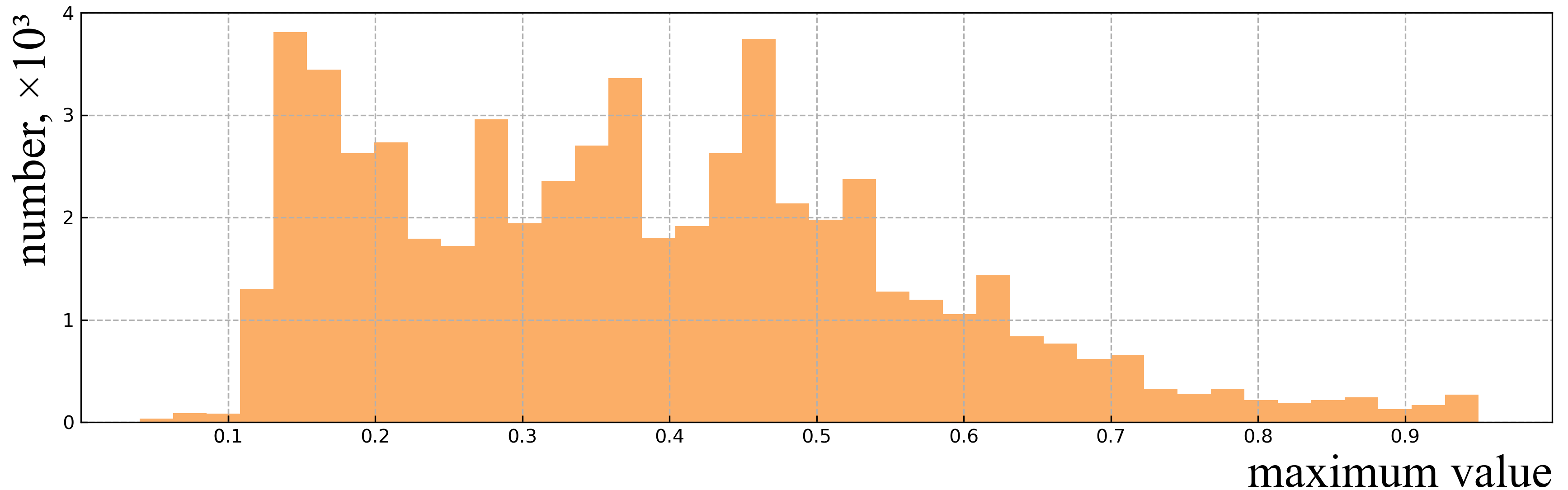}
  \vspace{-0.7em}
  \caption{The distribution result of the maximum value of $\boldsymbol{\alpha}$.}
  \label{Figs.hyper_parameter}
  \vspace{-1.2em}
  \end{figure}
\subsection{Ablation Study}
For efficiency, all ablation experiments are carried out by using SIGA$_{R}$.

\noindent \textbf{Network Structure.}
Our architecture consists of the Glyph Pseudo-label Construction (GPC), Glyph Attention Network (GLAN), and Attention-based Character Fusion Module (ACFM).
However, GLAN should be evaluated with GPC as a joint structure (JS) due to their interdependence, \emph{i.e.}, GLAN is supervised by glyph pseudo-labels generated from GPC during training. 
Thus we first perform the ``Baseline+JS'' model to evaluate the effectiveness of our structure. 
As depicted in Table \ref{tb:Ablation_Network}, 
the accuracy of ``Baseline+JS'' model is 7.01\% higher than ``Baseline''. Then, the gain of adding ACFM is further improved by 0.45\%, from 91.65\% to 92.10\% on average accuracy. 
Besides, we also encapsulate the JS structure as a plug-in component to SRN$\dagger$ and ABINet$\dagger$. As shown in the second group of rows in Table \ref{tb:Ablation_Network}, after using our JS structure, the average accuracy of these models is further improved by 5.68\% and 1.34\%, respectively.
These results demonstrate that the glyph attention extracted by the JS structure is effective and important to facilitate character recognition.

\noindent \textbf{Effectiveness of Implicit Attention Alignment.} We provide the effectiveness analysis and theoretical basis about our implicit attention alignment mudule in Supplementary Material.

\noindent \textbf{Hyper-parameters.}
The hyper-parameters map the attention weights $\boldsymbol{\alpha}$ to different numerical ranges.
For $\lambda $ and $\mu$ of Eq.\,\ref{eq1}, they encode $\boldsymbol{\alpha}$ to [0, 1] for generating $\boldsymbol{S}_{\rm sal}$ 
(Minimizing the difference between $\boldsymbol{S}_{\rm sal}$ and $\boldsymbol{S}_{\rm m}$ to alleviate the alignment drift problem).
For $\delta $ of Eq.\,\ref{eq2}, it binarizes $\boldsymbol{\alpha}$ to 0 or 1 for constructing $\boldsymbol{S}_{\rm gt}$ 
(Obtaining the glyph pseudo-labels).
To further observe the attention weight values $\boldsymbol{\alpha}$,  
we statistically analyze the distribution of the maximum value of $\boldsymbol{\alpha}$ on the validation set, as shown in Figure \ref{Figs.hyper_parameter}.
Thus the hyper-parameters are easily set to reasonable values based on the prior distribution.
Specifically, we set the confidence threshold $\delta $ to 0.05, 0.1, and 0.15, respectively, and the average accuracy on ten standard context benchmarks is 91.19\%, 90.75\%, and 90.43\%. 
For $\mu$ and $\lambda$ of Eq. \ref{eq1}, 
we set the three suitable parameter pairs ($\mu$, $\lambda$) to (100, 0.05), (70, 0.1), (40, 0.15), respectively, and the average accuracy is 91.02\%, 91.19\% and 90.83\%.

\begin{figure}[t]
  \centering
  \graphicspath{{./graph/}}
  \includegraphics[width=3.2in]{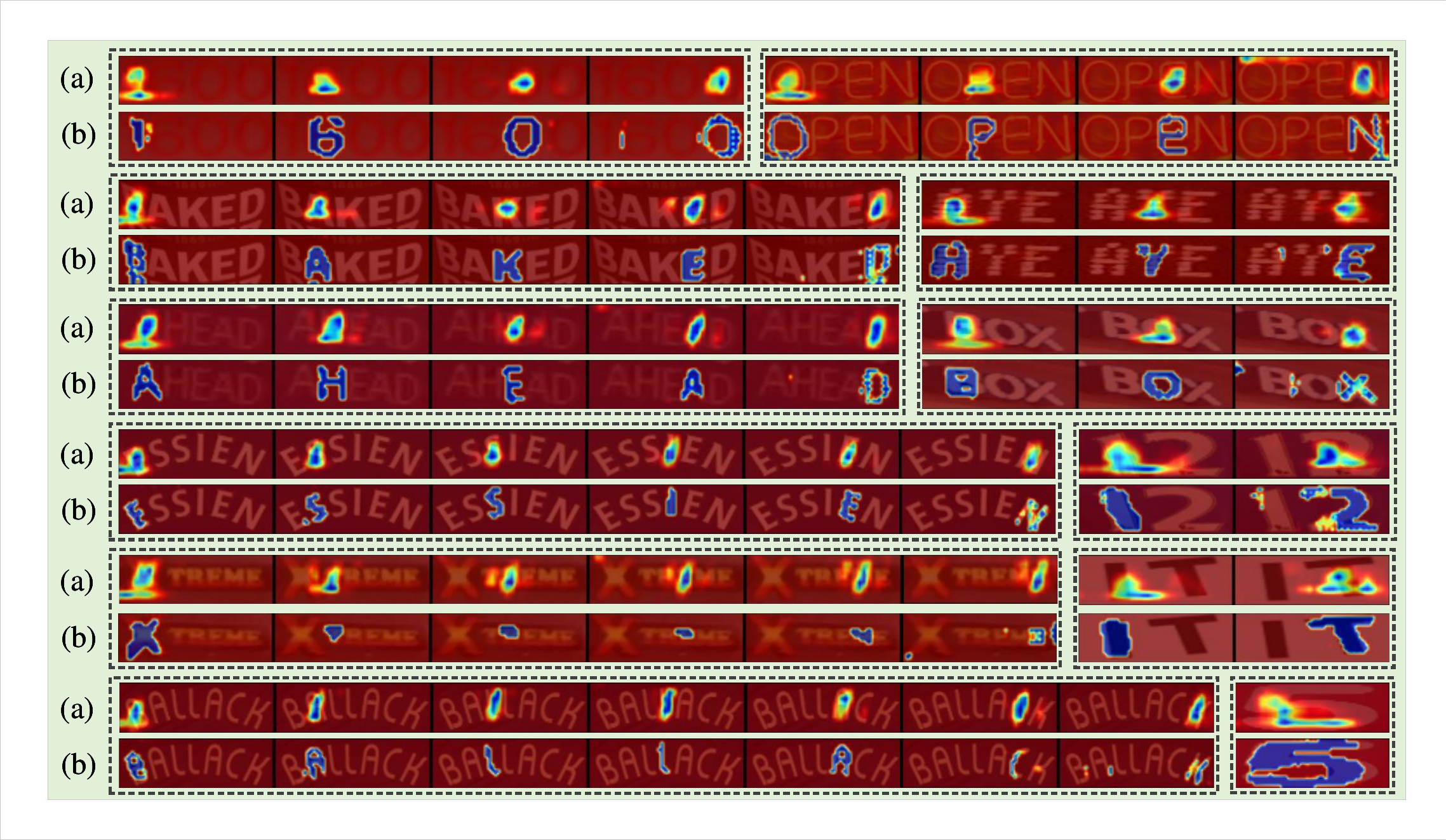}
  \vspace{-0.3em}
  \caption{Extensive 2D attention visualization results generated by ABINet (a) and SIGA (b). 
  }
  \label{Figs.Attention}
  \vspace{-1.2em}
\end{figure}  
\noindent \textbf{Visualization Analysis.} 
We have visualized the character attention maps in Figure \ref{Figs.Attention}, which indicates important regions contributed to character recognition. 
More visualization results of the SIGA method on horizontal, oriented, curved and blurred text images in Supplementary Material.
Specifically, each map in (a) and (b) is generated by the representative implicit attention method ABINet~\cite{ABINET:fang2021read} and our method SIGA, respectively. 
Unlike implicit attention mechanisms, our method perceives more fine-grained structural information of glyph.
Especially, our attention can still degenerate into the same attention form as other STR models in text images with very indistinct glyph (the fifth group of rows in Figure \ref{Figs.Attention}).
To the best of our knowledge, our method is the first to explore the glyph structures in STR. 

\noindent \textbf{Performance and Cost.}
We add speed and parameter comparisons in Table \ref{tb:Parameters_Speed}. 
For supervised attention methods (\emph{i.e.}, CA-FCN and TextScanner) with more parameters due to the large input size, \emph{e.g.}, 64$\times$256, SIGA$_{R}$ doesn't need extra character-level annotations and has better performance than their methods in Table \ref{tb:results}.
For implicit attention methods, SIGA$_{R}$ adds an unavoidable but acceptable amount of parameters (16.9M larger than ABINet$\dagger$, mainly used for a lightweight glyph attention structure) while obtaining more detailed glyph structures not being explored by other STR methods. 
Consequently, our method achieves the best performance in Table \ref{tb:results} and \ref{tb:contextless}.
\begin{table}[t]
  \centering
  \caption{Ablation study of the proposed SIGA$_{R}$ structure on context benchmarks. 
  ``JS'' means the joint structure of GPC and GLAN, as GLAN needs GPC for glyph pseudo-label construction. 
  } 
  \vspace{-0.5em}
  \scalebox{0.7}{
  \begin{tabular}{c|cccccc|c}
  \hline
  \multirow{2}{*}{\diagbox[width=9.5em]{Methods}{Datasets}} & IIIT & SVT & IC13 & IC15 & SP & CT &Average\\
      & 3000 & 647 & 857 & 1811 & 645 & 288 &7248\\ 
  \hline
  Baseline  &87.9 &87.5 &93.6 &77.6 &79.2 &74.0 &84.64  \\
  Baseline+JS  &95.7 &92.0 &96.6 &84.7 &85.7 &91.0 &91.65 \\
  Baseline+JS+ACFM &95.9 &92.7 &97.0 &85.1 &87.1 &91.7  &92.10 \\
  \hline
  SRN$\dagger$ &92.3 &88.1 &- &77.5 &79.4 &84.7 &86.04\\
  SRN$\dagger$+JS &96.2&92.7&96.3&84.3&85.4&90.0&91.72\\
  ABINet$\dagger $ &94.7 &91.7 &95.0 &82.7 &85.1 &86.5 &90.29\\
  ABINet$\dagger $+JS &95.9 &92.0 &96.4 &84.4 &85.6 &91.0 &91.63  \\
  \hline
  \end{tabular}
  }
  \label{tb:Ablation_Network}
\end{table}
\begin{table}[t]
  \centering
  \caption{Comparison results of speed and parameter amount.}
  \vspace{-0.5em}
  \scalebox{0.7}{
  \begin{tabular}{c|c|c|c|c}
  \toprule
  Methods & Description & Size &Param. (MB)&Time (ms)\\
  \midrule
  CA-FCN~\cite{liao2019scene} &Sup.&${64\times 256}$ &-&-\\
  TextScanner~\cite{wan2020textscanner} &Sup.&${64\times 256}$ &57&56.8\\
  SRN$\dagger $~\cite{SRN:yu2020towards} &Implicit&${64\times256}$ &41.4&131.5\\
  TRBA~\cite{TRBA:baek2019wrong}&Implicit &${32\times100}$  &49.6&27.6\\
  ABINet$\dagger$~\cite{ABINET:fang2021read} &Implicit&${32\times128}$ &23.5&16.7\\
  SIGA$_{R}$ &Self-Sup.&${32\times128}$  &40.4&53.7\\
  \midrule
  CornerTransformer~\cite{xie2022toward}&Implicit&${32\times128}$ &85.7&294.9\\
  LevOCR~\cite{LevOCR} &Implicit&${32\times128}$ &109.0&119.0\\
  MGP-STR~\cite{MGP}&Implicit&${32\times128}$  &148.0&12.3\\
  SIGA$_{T}$&Self-Sup.&${32\times128}$  &113.1&56.3\\
  \bottomrule    
  \end{tabular}
  }
  \vspace{-1.0em}
  \label{tb:Parameters_Speed}
  \end{table}
\section{Conclusions}
In this paper, we propose a novel attention-based method for STR, Self-supervised Implicit Glyph Attention (SIGA). Beyond the difficulty of character-level annotation by humans, SIGA delineates the glyph structures of text images as the supervision of attention maps by jointly self-supervised text segmentation and implicit attention alignment.
The learned glyph attention then encourages the
text recognition network to focus on the structural regions of glyphs to improve attention correctness.
Finally, extensive experiments demonstrate that SIGA achieves the best performance on context and contextless benchmarks. 

\noindent \textbf{Acknowledgements} 
This work was supported by National Key Research and Development Project of China under Grant 2019YFB1706602, NSFC 62273235, NSFC 62176159, Natural Science Foundation of Shanghai 21ZR1432200, Shanghai Municipal Science and Technology Major Project 2021SHZDZX0102 and the Fundamental Research Funds for the Central Universities. 

\newpage
\appendix
\title{Self-supervised Implicit Glyph Attention for Text Recognition\\(Supplementary Material)}

\author{Tongkun Guan\textsuperscript{\rm 1}, Chaochen Gu\textsuperscript{\rm 2\footnotemark[1]},
Jingzheng Tu\textsuperscript{\rm 2},
Xue Yang\textsuperscript{\rm 1},
Qi Feng\textsuperscript{\rm 2},
Yudi Zhao\textsuperscript{\rm 2},\\
Xiaokang Yang\textsuperscript{\rm 1},
Wei Shen\textsuperscript{\rm 1\footnotemark[1]}\\
\textsuperscript{\rm 1} MoE Key Lab of Artificial Intelligence, AI Institute, Shanghai Jiao Tong University\\
\textsuperscript{\rm 2} Department of Automation, Shanghai Jiao Tong University\\
{\tt\small \{gtk0615,jacygu,wei.shen\}@sjtu.edu.cn}
}
\maketitle
\vspace{-0.5em}
\section{Further Details for Text Datasets}
\begin{figure}[t]
  \centering
  \includegraphics[width=3.2in]{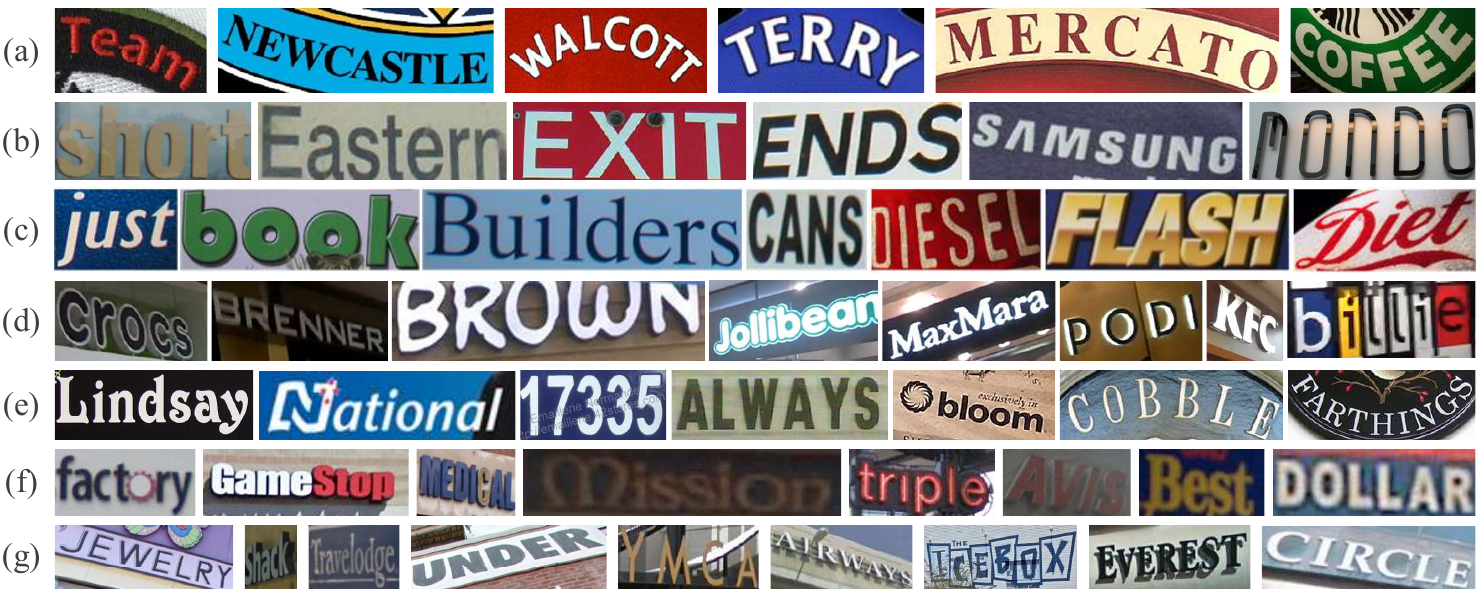}
  \vspace{-0.8em}
  \caption{Some examples of natural scene text datasets. \\(a) CUTE80\cite{CT:risnumawan2014robust}; (b) ICDAR2003\cite{lucas2005icdar}; (c) ICDAR2013\cite{IC13:karatzas2013icdar}; (d) ICDAR2015\cite{IC15:karatzas2015icdar}; (e) IIIT5k\cite{IIIT5K:mishra2012scene}; (f) SVT\cite{SVT:wang2011end}; (g) SVTP\cite{SP:phan2013recognizing}.} 
  \vspace{-1.0em}
  \label{Figs.STD_benchmarks}
  \end{figure}

\vspace{-0.5em}
In this section, we present more visualizations of text datasets. 
As shown in Figure\,\ref{Figs.STD_benchmarks}, the existing scene text recognition datasets are taken from natural scenes, including traffic signs, shopping mall trademarks, billboards, \emph{etc}. These images have relatively clear texts with variable styles and colours against a chaotic background. 

\begin{figure}[t]
  \centering
  \includegraphics[width=3.2in]{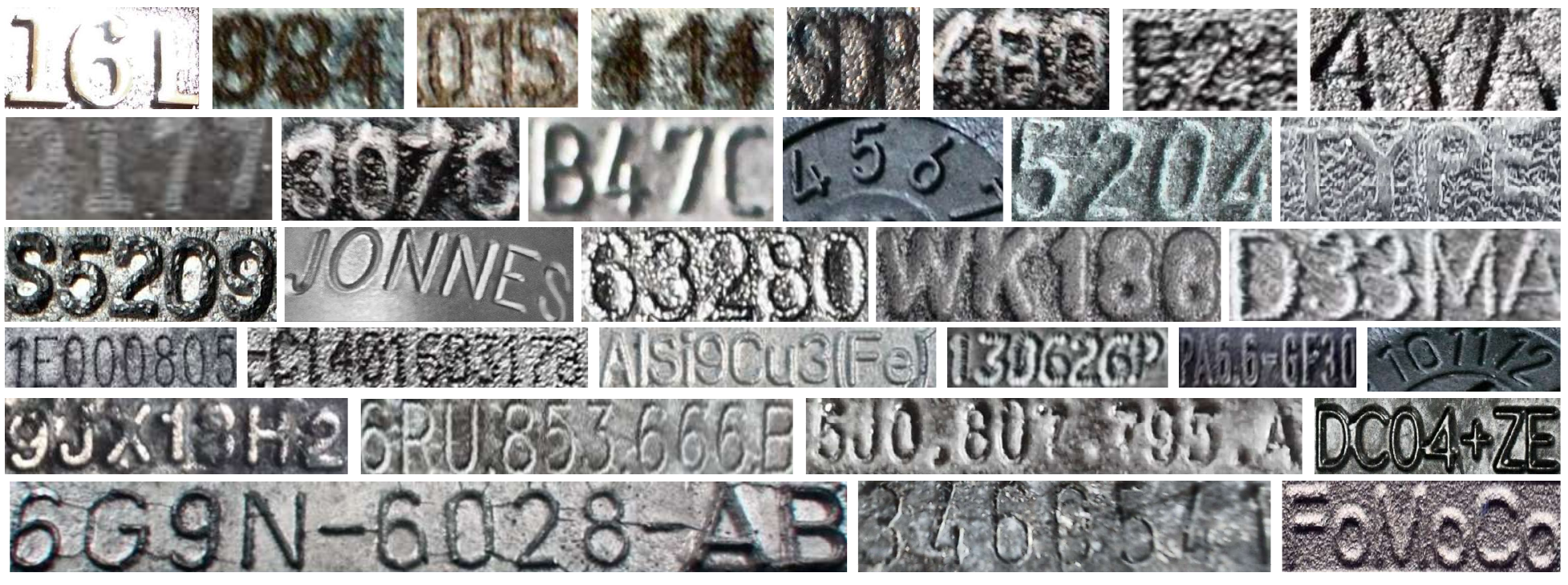}
  \vspace{-0.8em}
  \caption{Some examples of MPSC dataset.}
  \vspace{-1.0em}
  \label{Figs.MPSC}
  \end{figure}

\begin{figure}[t]
  \centering
  \includegraphics[width=3.2in]{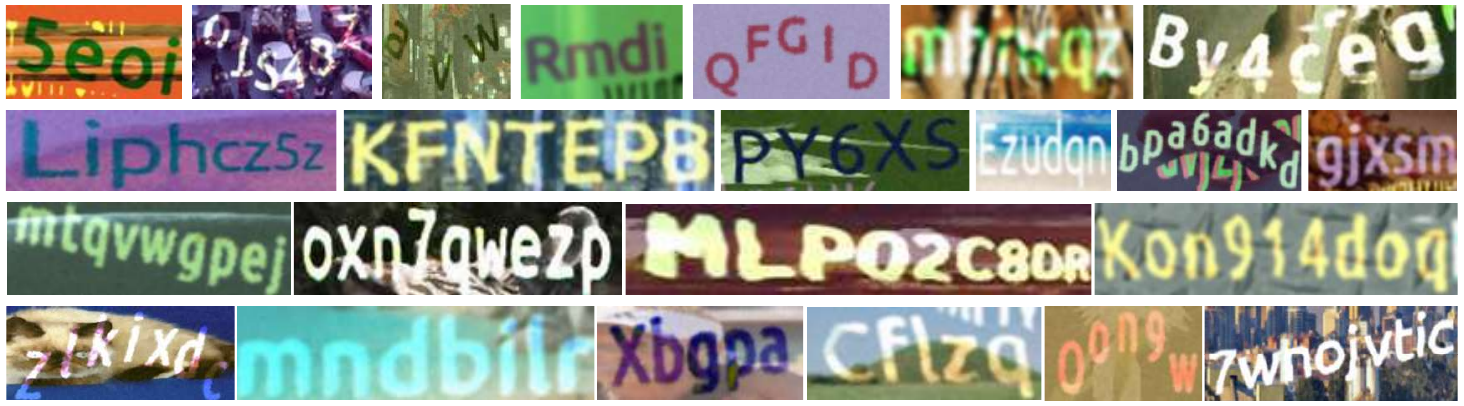}
  \vspace{-0.8em}
  \caption{Some examples of ArbitText dataset.}
  \vspace{-1.0em}
  \label{Figs.ArbitText}
  \end{figure}

In contrast, the MPSC dataset contains many contextless texts with low visual contrast, corroded surfaces, and uneven illumination as shown in Figure\,\ref{Figs.MPSC}, which poses a new challenge to contextless text recognition. Specifically, these text images are marked with Latin characters and Arabic numerals to record the serial number, production date, and other product information. Recognizing these texts plays an increasingly important role in intelligent industrial manufacturing, which is conducive to improving the assembly speed of industrial production lines and the efficiency of logistics transmission in the industrial scene. Besides, as shown in Figure\,\ref{Figs.ArbitText}, we employ the synthetic tool~\cite{yim2021synthtiger} by selecting the appropriate background images and various fonts and colours to generate these text images. Each text of the ArbitText dataset contains a random combination of Latin characters and Arabic numerals. The whole dataset contains 1M images, which is used to evaluate the generalizability and efficiency of language-free models on contextless texts. 
\vspace{-1.0em}
\section{Effectiveness of IAA Module}
\vspace{-0.5em}
We measure the effects of our implicit attention alignment (IAA) module on the finely annotated dataset, TextSeg.
  \begin{figure*}[t]
    \vspace{-1.0em}
    \centering
    \includegraphics[width=6.8in]{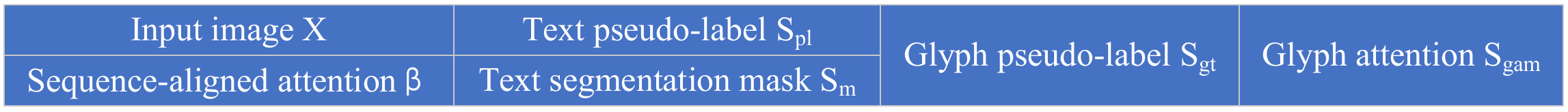}
    \vspace{-1.0em}
    \caption{The arrangement order.}
    \vspace{-1.0em}
    \label{Figs.arrangement}
    \end{figure*}
    \begin{figure*}[t]
      \centering
      \includegraphics[width=6.8in]{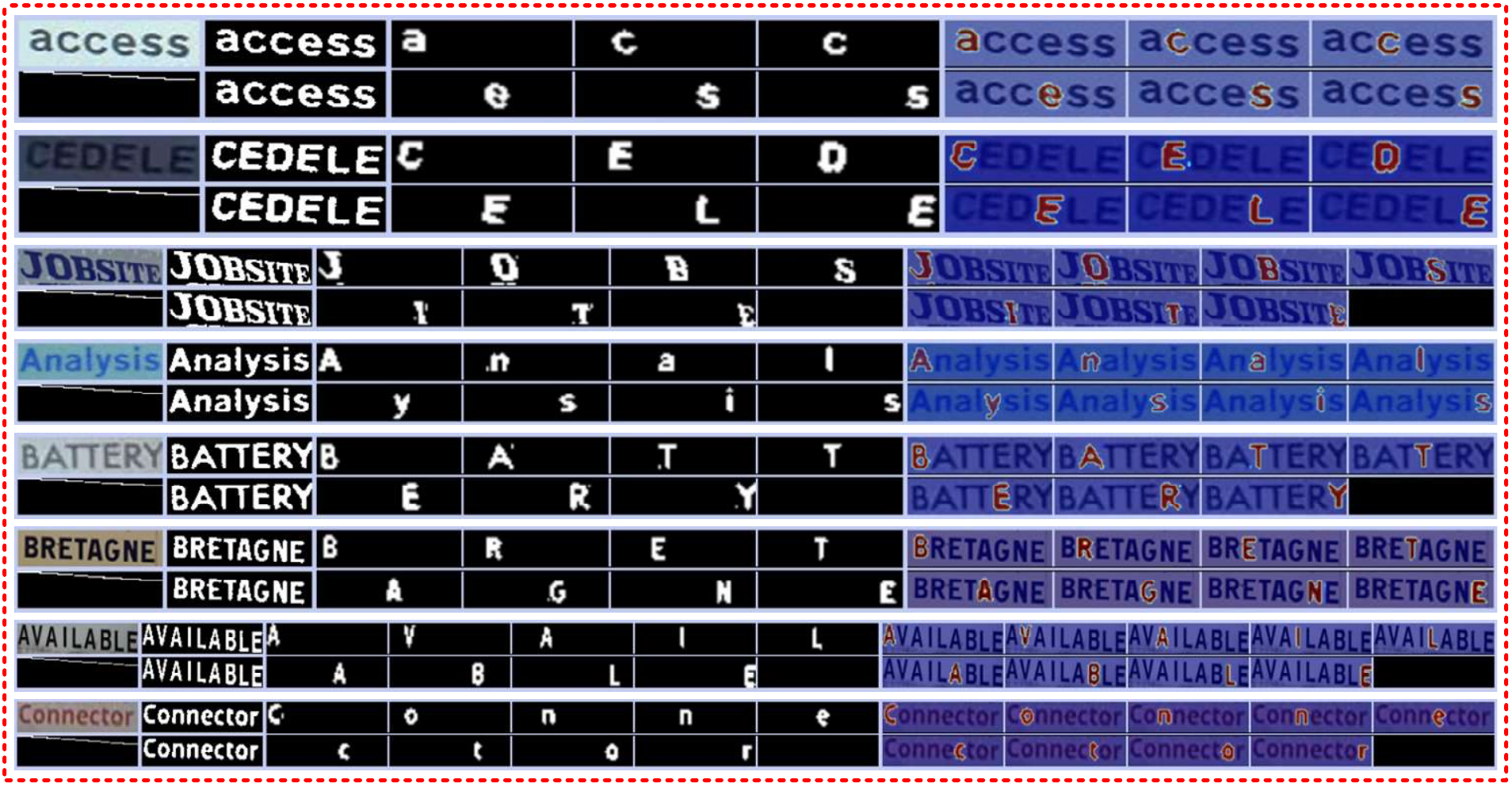}
      \vspace{-0.5em}
      \caption{Visualization results of the SIGA method on horizontal text images.}
      \vspace{-1.0em}
      \label{Figs.h}
    \end{figure*}
\vspace{-0.3em}
\subsection{Metric}
\vspace{-0.3em}
Let $\boldsymbol{b} \in \{0, 1\}^{H\times W}$ be the character mask generated by assigning 1 to the locations in the ground-truth character box and 0 otherwise, we calculate its horizontal projections $\boldsymbol{l} \in \{0, 1\}^{W}$ by a $\rm max$ operation of $\boldsymbol{b}$ along with $x$-axis. We then assume that $\boldsymbol{\tilde{l}} \in \{0, 1\}^{W}$ denotes the thresholded network predictions ($>0.05=1$) for the attention of corresponding character, the metric $\Theta$ is defined as: $\Theta = \boldsymbol{l} \cdot \boldsymbol{\tilde{l}} / \Vert \boldsymbol{l} + \boldsymbol{\tilde{l}} -\boldsymbol{l} \cdot \boldsymbol{\tilde{l}} \Vert_1$. And then, we also evaluate their average recognition accuracies on the ten standard context benchmarks.
Specifically, the detailed ablation results are as illustrated in Table \ref{tb:iaa}.
\begin{table}[t]
  \centering
  \caption{Ablation results of different loss components.}
  \vspace{-0.5em}
  \scalebox{0.8}{
  \begin{tabular}{|c|c|c|c|c|}
  \hline
       Loss & - & $\mathcal{L}_{\rm cor}$ & $\mathcal{L}_{\rm dif}$ &$\mathcal{L}_{\rm cor} + \mathcal{L}_{\rm dif}$ \\
       \hline
       $\Theta$\%(ACC\%) & 53.2(69.1) & 55.2(69.4) & 60.5(70.0) & 63.6(70.5)\\
       \hline
  \end{tabular}
  }
  \label{tb:iaa}
  \vspace{-1.0em}
\end{table}
\subsection{Theoretical Basis}
Given a normalized image, let $\boldsymbol{l}_{t}, \tilde{\boldsymbol{l}_{t}} \in \{0, 1\}^{W}$ be its ground-truth horizontal projections and the thresholded network predictions for the attention at the decoding time $t$, we target on $\tilde{\boldsymbol{l}_{t}} = \boldsymbol{l}_{t}, \forall t \in \{1,...,T\}$ to mitigate the alignment drift issue. Specifically, we propose a constraint function in implicit attention alignment module, which can be summarized as follows:
\begin{equation}
    \begin{aligned}
      \sum_{1\leqslant i \textless j \leqslant T} \Tilde{\boldsymbol{l}}_{i} \cdot \Tilde{\boldsymbol{l}}_{j} \to 0,
      \sum_{i=1}^{T}(\psi(\Tilde{\boldsymbol{l}}_{i})\cdot\Tilde{\boldsymbol{M}}) \to \Tilde{\boldsymbol{M}},
    \end{aligned}
    \end{equation}
    
\noindent where $\Tilde{\boldsymbol{M}} \in (0,1)^{W \times H}$ is our  network predictions for text mask and $\psi : \mathbb{R}^{W} \to \mathbb{R}^{W \times H}$ with a dimension expansion.

Ideally, define $\boldsymbol{M}$ as the ground-truth text mask, suppose $\Tilde{\boldsymbol{M}} = \boldsymbol{M}$, the target $\tilde{\boldsymbol{l}_{t}} = \boldsymbol{l}_{t}, \forall t \in \{1,...,T\}$ is a good feasible solution as: 
\begin{equation}
    \begin{aligned}
      \textstyle{\sum_{1\leqslant i \textless j \leqslant T} \boldsymbol{l}_{i} \cdot \boldsymbol{l}_{j} = 0,
      \sum_{i=1}^{T}(\psi(\boldsymbol{l}_{i})\cdot\boldsymbol{M}) = \boldsymbol{M}}
    \end{aligned}
\end{equation}

Although the target is a necessary but not sufficient condition for our constraint function as some extreme cases exist, the generality where the attention mechanism works in most images, ensures that SIGA can toward the target, which is also demonstrated by the above-mentioned ablation results. 

\section{Visualizations of Glyph Attention}
In SIGA, five important items assist the text recognition network to obtain glyph features for improving performance. 
They are text pseudo-label $\boldsymbol{S}_{pl}$, sequence-aligned weights $\boldsymbol{\beta}$, text segmentation mask $\boldsymbol{S}_{m}$, glyph pseudo-labels $\boldsymbol{S}_{gt}$, and glyph attention maps $\boldsymbol{S}_{gam}$, respectively. 

Specifically, given an input image X, SIGA first employs the $K$-means algorithm to generate a text pseudo-label $\boldsymbol{S}_{pl}$, and further utilizes the text pseudo-label to optimize our designed self-supervised text segmentation module to generate a text segmentation mask $\boldsymbol{S}_{m}$. Then, we follow an implicit attention method as the baseline structure to obtain implicit attention weights $\boldsymbol{\alpha}$, which are transformed into sequence-aligned attention vectors $\boldsymbol{\beta}$ by an orthogonal constraint, and served as the position information of characters in the input image X. Next, we obtain the glyph pseudo-label $\boldsymbol{S}_{pl}$ via the dot product operation between the sequence-aligned attention vectors $\boldsymbol{\beta}$ and the learned text segmentation mask $\boldsymbol{S}_{m}$. Finally, supervised by the glyph pseudo-label $\boldsymbol{S}_{pl}$, our text recognition network produces glyph attention maps $\boldsymbol{S}_{gam}$.

\begin{figure*}[!t]
  \centering
  \includegraphics[width=6.8in]{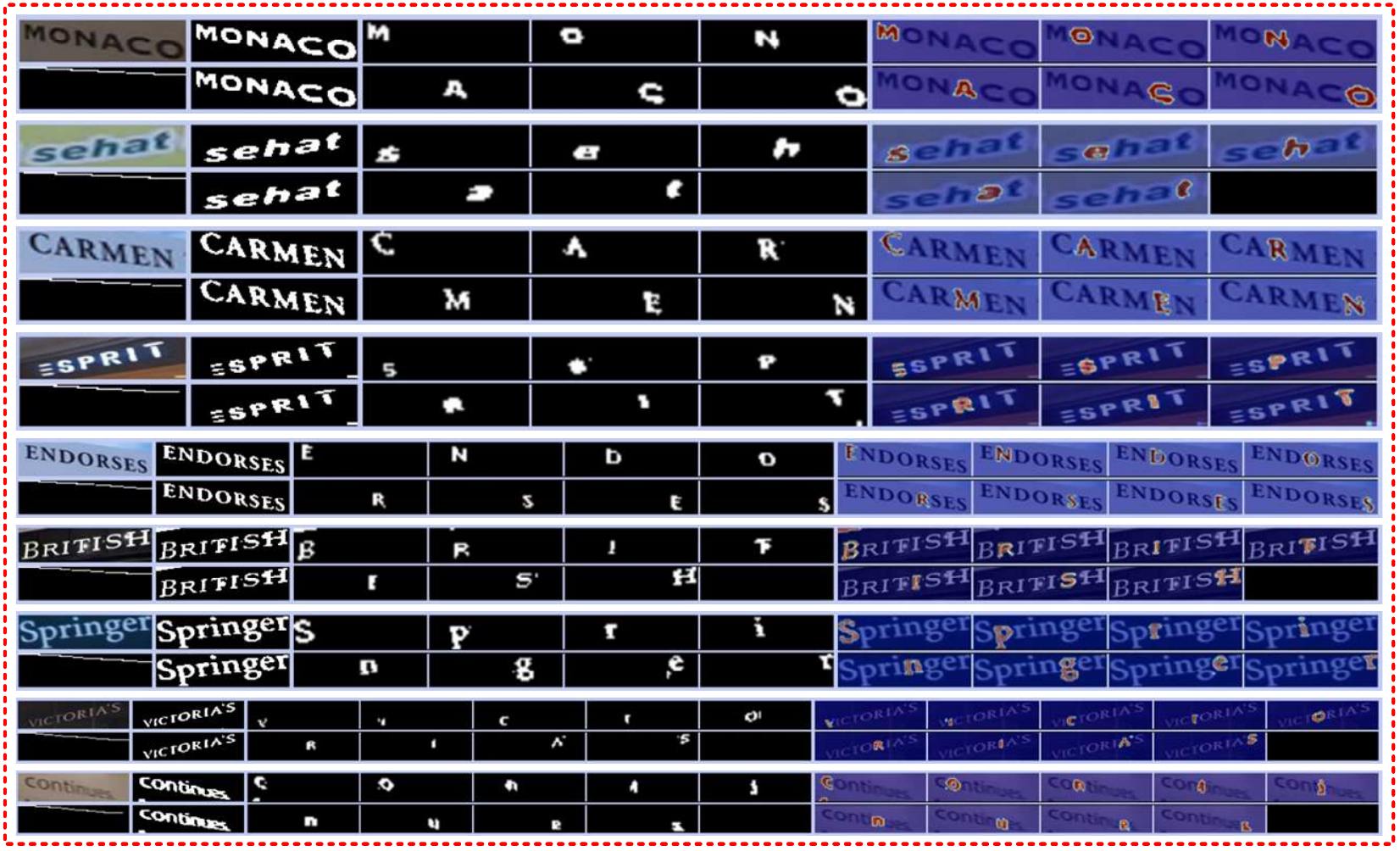}
  \vspace{-0.5em}
  \caption{Visualization results of the SIGA method on oriented text images.}
  \label{Figs.r}
\end{figure*}

\begin{figure*}[!t]
  \centering
  \includegraphics[width=6.8in]{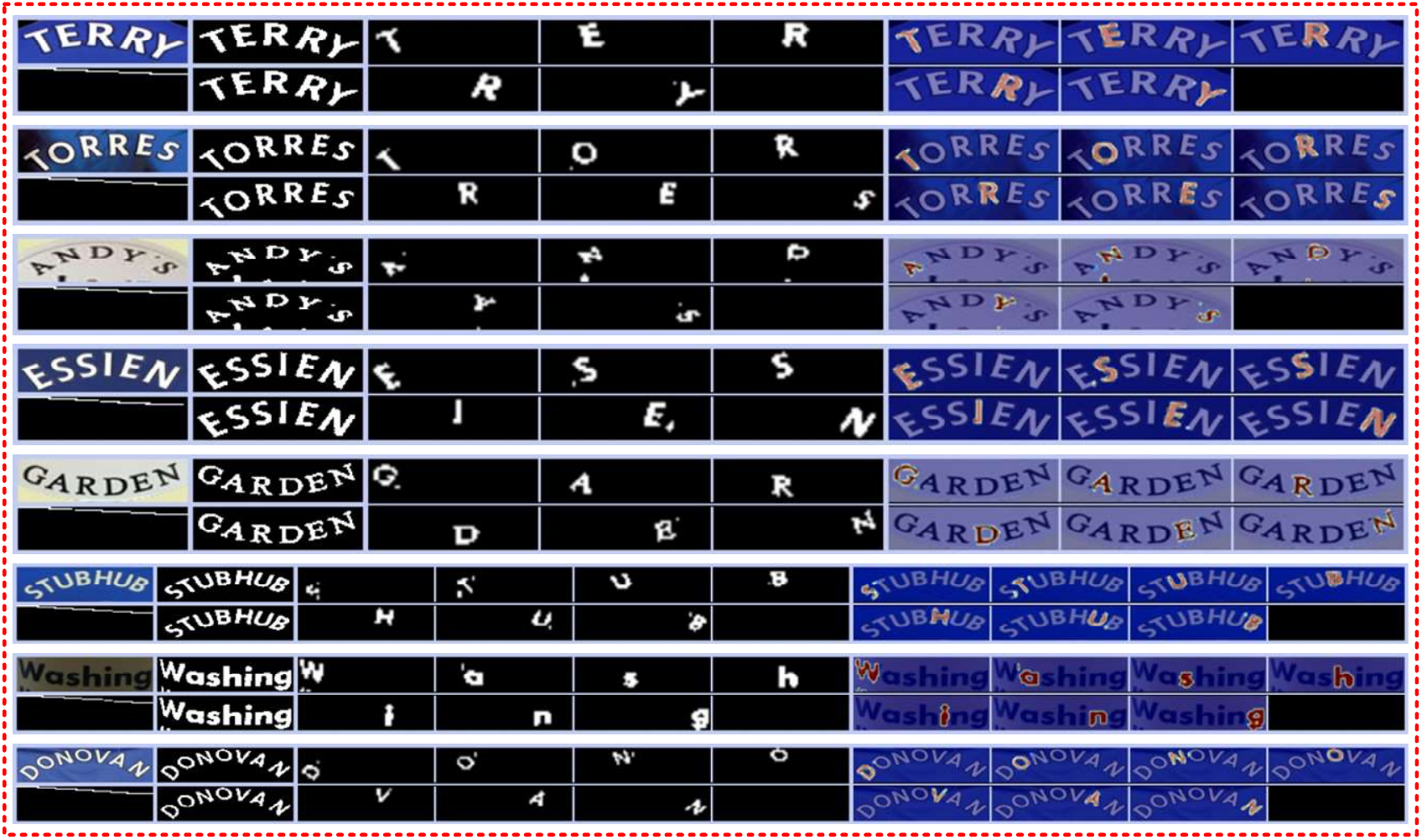}
  \vspace{-0.8em}
  \caption{Visualization results of the SIGA method on curved text images.}
  \label{Figs.c}
\end{figure*}

\begin{figure*}[!t]
  \vspace{-0.8em}
  \centering
  \includegraphics[width=\textwidth]{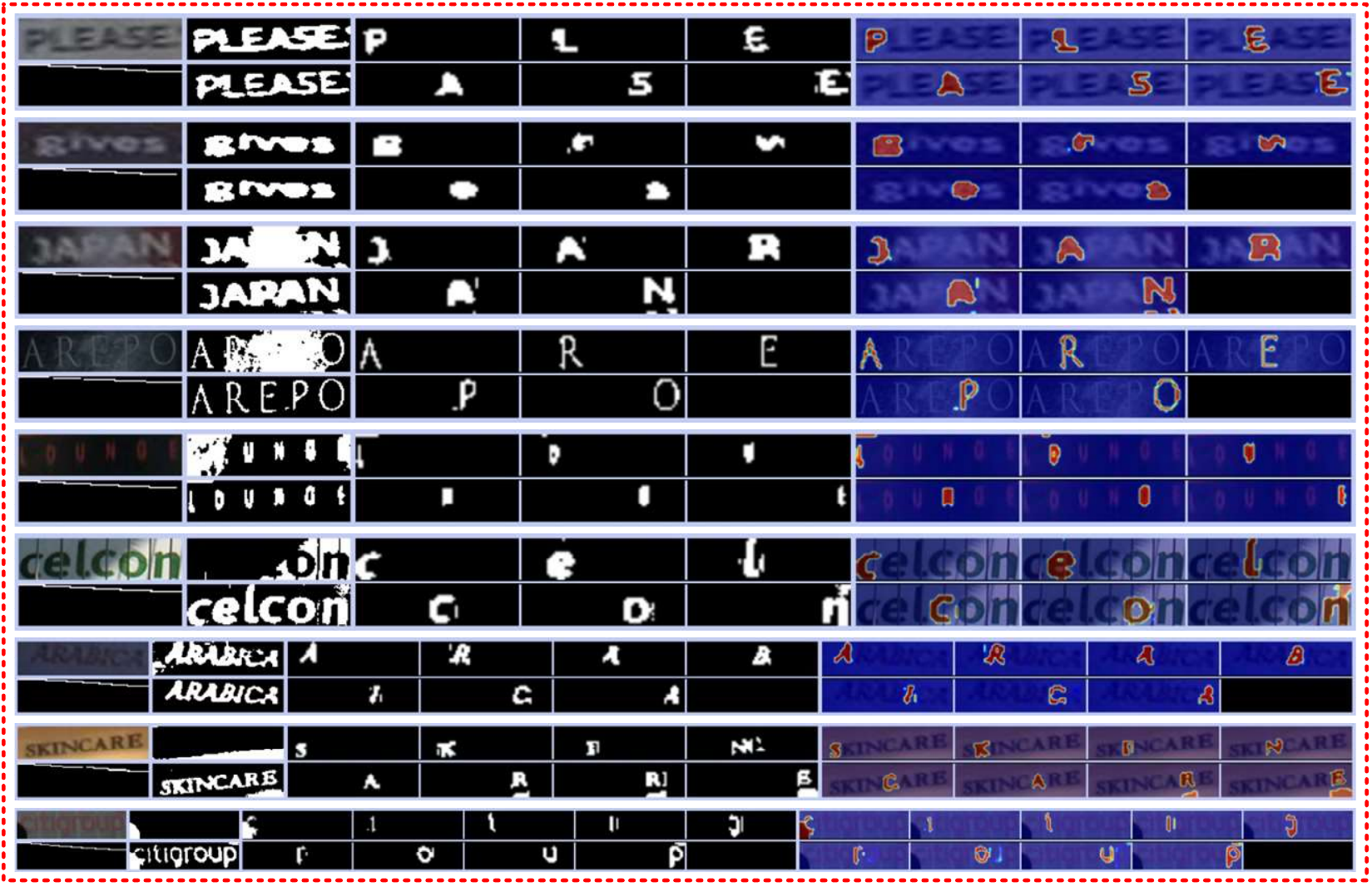}
  \vspace{-1.8em}
  \caption{Visualization results of the SIGA method on blurred text images.}
  \label{Figs.b}
\end{figure*}

To further illustrate the generation pipeline of glyph structures in SIGA, as shown in Figure\,\ref{Figs.h}-\ref{Figs.b}, we visualize more examples of these items on horizontal, oriented, curved, and blurred text images. Specifically, every example follows the arrangement order in Figure\,\ref{Figs.arrangement}. 

{\small
\bibliographystyle{ieee_fullname}
\bibliography{egbib}
}
\end{document}